\documentclass[journal]{IEEEtran}
\ifCLASSINFOpdf
\else
\fi

\makeatletter

\newcommand{\Rmnum}[1]{\expandafter\@slowromancap\romannumeral #1@}
\makeatother
\usepackage{times}
\usepackage{epsfig}
\usepackage{graphicx}
\usepackage{amsmath}
\usepackage{amssymb}
\usepackage{multirow}
\usepackage{tabularx}
\usepackage{caption}
\usepackage{subcaption}
\usepackage{bm}
\usepackage{booktabs}
\usepackage{float}
\usepackage{array}
\usepackage{cite}
\usepackage{epstopdf}
\usepackage{slashbox}
\usepackage[dvipsnames]{xcolor}
\usepackage{hyperref}
\usepackage{cleveref}
\usepackage{indentfirst}
\usepackage{color}
\usepackage{threeparttable}
\newcommand{\tabincell}[2]{\begin{tabular}{@{}#1@{}}#2\end{tabular}}

\ifCLASSOPTIONcompsoc
  \usepackage[nocompress]{cite}
\else
  \usepackage{cite}
\fi
\newcommand\myshade{85}
\colorlet{mylinkcolor}{violet}
\colorlet{mycitecolor}{purple}%yelloworange
\colorlet{myurlcolor}{Aquamarine}

\hypersetup{
  linkcolor  = mylinkcolor!\myshade!black,
  citecolor  = mycitecolor!\myshade!black,
  urlcolor   = myurlcolor!\myshade!black,
  colorlinks = true,
}

\begin{document}
\title{Adversarial Transfer Learning for Cross-domain Visual Recognition}

\author{Shanshan Wang,~\IEEEmembership{Student Member,~IEEE,~}
        Lei Zhang,~\IEEEmembership{Senior Member,~IEEE,~}
        Jingru Fu,~\IEEEmembership{Student Member,~IEEE~}
      %  Wangmeng Zuo,~\IEEEmembership{Senior Member,~IEEE,~}% <-this % stops a space
%        Bob Zhang,~\IEEEmembership{Member,~IEEE~}\\% <-this % stops a space
\IEEEcompsocitemizethanks{\IEEEcompsocthanksitem  This work was supported by the National Science Fund of China under Grants (61771079) and the Chongqing Natural Science Fund (No. cstc2018jcyjAX0250). (\textit{Corresponding author: Lei Zhang})

S. Wang, L. Zhang  and J. Fu are with the School of Microelectronics and Communication Engineering, Chongqing University, Chongqing 400044, China.
\ (E-mail: wangshanshan@cqu.edu.cn, leizhang@cqu.edu.cn, jrfu@cqu.edu.cn).

%\thanks{This work was supported by the National Science Fund of China under Grants (61401048, 91420201 and 61472187), the Fundamental Research Funds for the Central Universities (No. 106112017CDJQJ168819), the 973 Program No.2014CB349303, and Program for Changjiang Scholars.
}}


\markboth{IEEE TRANSACTIONS ON XXXX,~Vol.~-, No.~-, January~2019}%
{Shell \MakeLowercase{\textit{et al.}}: Bare Demo of IEEEtran.cls for IEEE Journals}



\maketitle

\begin{abstract}
 In many practical visual recognition scenarios, feature distribution in the source domain is generally different from that of the target domain, which results in the emergence of general cross-domain visual recognition problems. To address the problems of visual domain mismatch, we propose a novel semi-supervised adversarial transfer learning approach, which is called \textbf{C}oupled \textbf{a}dversarial \textbf{t}ransfer \textbf{D}omain \textbf{A}daptation (CatDA), for distribution alignment between two domains. The proposed CatDA approach is inspired by cycleGAN, but leveraging multiple shallow multilayer perceptrons (MLPs) instead of deep networks. Specifically, our CatDA comprises of two symmetric and slim sub-networks, such that the coupled adversarial learning framework is formulated. With such symmetry of two generators, the input data from source/target domain can be fed into the MLP network for target/source domain generation, supervised by two confrontation oriented coupled discriminators. Notably, in order to avoid the critical flaw of high-capacity of the feature extraction function during domain adversarial training, domain specific loss and domain knowledge fidelity loss are proposed in each generator, such that the effectiveness of the proposed transfer network is guaranteed. Additionally, the essential difference from cycleGAN is that our method aims to generate domain-agnostic and aligned features for domain adaptation and transfer learning rather than synthesize realistic images. We show experimentally on a number of benchmark datasets and the proposed approach achieves competitive performance over state-of-the-art domain adaptation and transfer learning approaches.
\end{abstract}


\section{Introduction}

\begin{figure}[t]
\centering
 \includegraphics[width=0.13\linewidth,height=1.2cm]{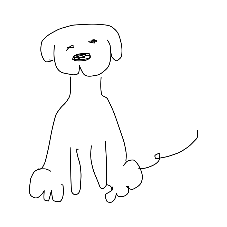}
  \includegraphics[width=0.13\linewidth,height=1.2cm]{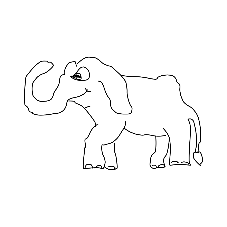}
  \includegraphics[width=0.13\linewidth,height=1.2cm]{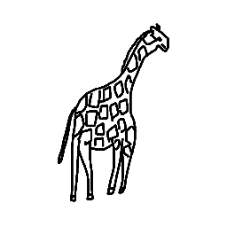}
  \includegraphics[width=0.13\linewidth,height=1.2cm]{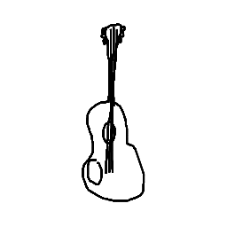}
  \includegraphics[width=0.13\linewidth,height=1.2cm]{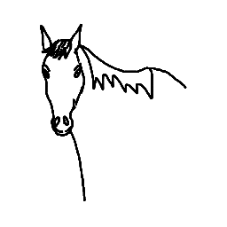}
  \includegraphics[width=0.13\linewidth,height=1.2cm]{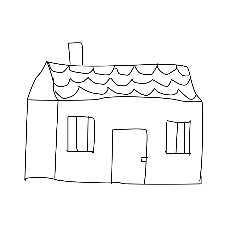}
  \includegraphics[width=0.13\linewidth,height=1.2cm]{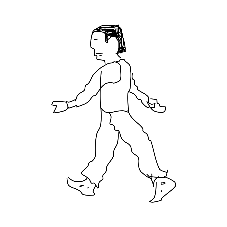}
  \includegraphics[width=0.13\linewidth,height=1.2cm]{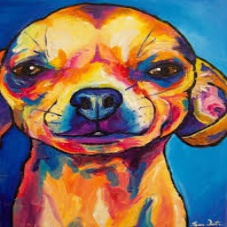}
  \includegraphics[width=0.13\linewidth,height=1.2cm]{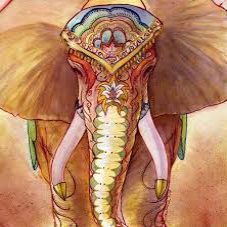}
  \includegraphics[width=0.13\linewidth,height=1.2cm]{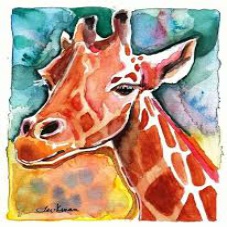}
  \includegraphics[width=0.13\linewidth,height=1.2cm]{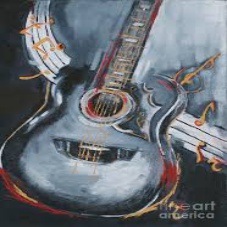}
  \includegraphics[width=0.13\linewidth,height=1.2cm]{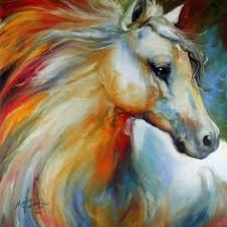}
  \includegraphics[width=0.13\linewidth,height=1.2cm]{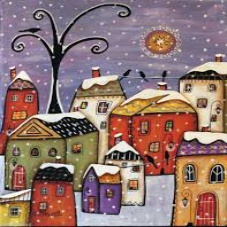}
  \includegraphics[width=0.13\linewidth,height=1.2cm]{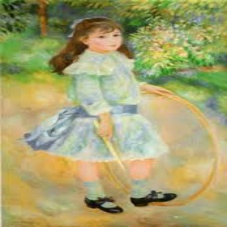}
  \includegraphics[width=0.13\linewidth,height=1.2cm]{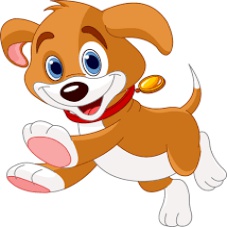}
  \includegraphics[width=0.13\linewidth,height=1.2cm]{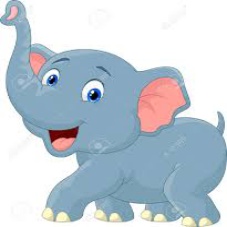}
  \includegraphics[width=0.13\linewidth,height=1.2cm]{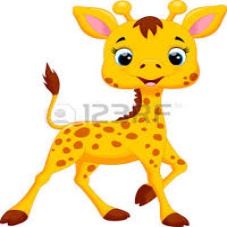}
  \includegraphics[width=0.13\linewidth,height=1.2cm]{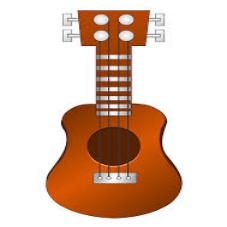}
  \includegraphics[width=0.13\linewidth,height=1.2cm]{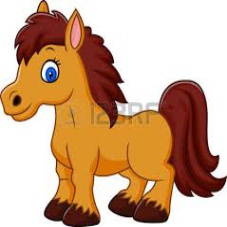}
  \includegraphics[width=0.13\linewidth,height=1.2cm]{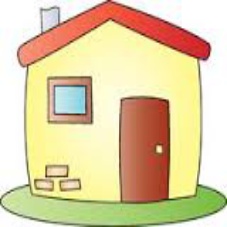}
  \includegraphics[width=0.13\linewidth,height=1.2cm]{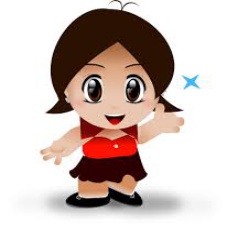}
    \includegraphics[width=0.13\linewidth,height=1.2cm]{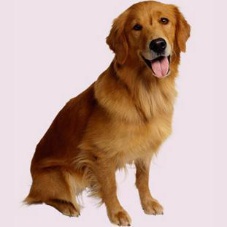}
  \includegraphics[width=0.13\linewidth,height=1.2cm]{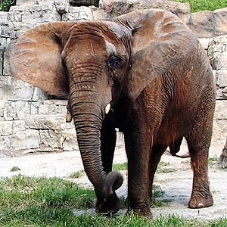}
  \includegraphics[width=0.13\linewidth,height=1.2cm]{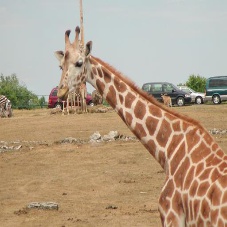}
  \includegraphics[width=0.13\linewidth,height=1.2cm]{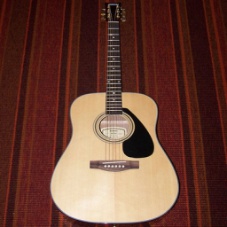}
  \includegraphics[width=0.13\linewidth,height=1.2cm]{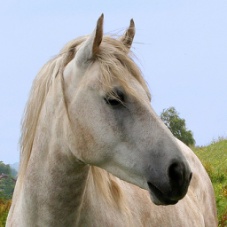}
  \includegraphics[width=0.13\linewidth,height=1.2cm]{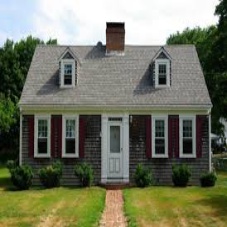}
  \includegraphics[width=0.13\linewidth,height=1.2cm]{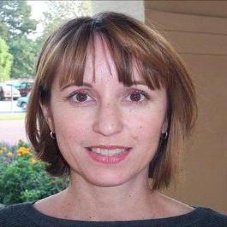}
   \caption{Different distributions from different domains. It is not difficult to understand that if the images in the first row are used to train an effective classifier, the model cannot work well for classifying the remaining images.}
   \label{figdatasets}
\end{figure}

\IEEEPARstart{I}{n} cross-domain visual recognition systems, the traditional (task-specific) classifiers usually do not work well on those semantic-related but distribution different tasks. {A typical cross-domain problem is presented in Fig.~\ref{figdatasets}, which illustrates some example images with similar semantics but different distribution}. By observing the Fig.~\ref{figdatasets}, {it is not difficult to understand that the classifier trained with the images in the first row cannot work well when classifying the remaining images due to the explicit heterogeneity of multiple visual tasks}. Mathematically, the reason lies in that the training data and test samples have different feature distribution (i.e. data bias) and do not satisfy the condition of independent identical distribution (i.e. i.i.d.)~\cite{pan2010survey,Lu2017When,Zhang2017Odor,Turner2017Transfer,Yan2017Drift}. Additionally, to train an accurate classification model, sufficient labeled samples are needed according to the statistical machine learning theory. However, collecting data is an expensive and time-consuming process. Generally, the data problem can be relieved by exploiting a few label information from related source domains and leveraging a number of unlabeled instances from the target domain for recognition model training. {Although the scarcity of data and labels can be partially solved by fusing the data drawn from multiple different domains, another dilemma of domain discrepancy is resulted.} Recently, domain adaptation (DA)~\cite{Duan2010Visual,kulis2011you,Zhang2015Odor} techniques which can effectively ease domain shift problem have been proposed, and demonstrated a great success in various cross-domain visual datasets.

It is of great practical importance to explore DA methods. DA models allow machine learning model to be self-adapted among multiple visual knowledge domains, i.e., the trained model from one data domain can be adapted to another domain and it is the key objective of this paper. There is a fundamental assumption underlying is that, although the domains differ, there is sufficient commonality to support such adaptation.

Most of the existing DA algorithms seek to bridge the gap between domains by reconstructing a common feature subspace for general feature representation. In this paper, we reformulate DA as a conditional image generation problem which tends to bridge the gap by generating domain specific data. The mapping function from one domain to another can be viewed as the modeling process of a generator, which achieves automatic domain shift alignment during data sampling and generating \cite{Kim2017Learning}. Recently, generative adversarial network (GAN) proposed in~\cite{Goodfellow2014Generative}, that tends to generate user-defined images by the adversarial mechanism between generator and discriminator, has become a mainstream of DA approach~\cite{ganin2017domain}. This is usually modeled by minimizing the approximate domain discrepancy via an adversarial objective function. GAN generally carries two networks called generator and discriminator, which work against each other. The generator is trained to produce images that could confuse the discriminator, while the discriminator tries to distinguish the generated images from real images. This adversarial strategy is very suitable for DA problem~\cite{Tzeng2017Adversarial,Salimans2016Improved}, because domain discrepancy can be easily reduced by adversarial generation. Therefore, this confrontation principle is exploited to ensure that the discriminator cannot distinguish the source domain from the generated target domain. In~\cite{ganin2017domain}, DANN is one of the first work of deep domain adaptation, in which the adversarial mechanism of GAN was used to bridge the gap between the source and target domains.
Similarly, the GAN inspired domain adaptation (ADDA) with convolutional neural network (CNN) architecture has also achieved a surprisingly good performance in~\cite{Tzeng2017Adversarial}. With the success of GAN in domain adaptation, an adversarial domain adaptation framework with domain generators and domain discriminators as GAN does is studied in this work for cross-domain visual recognition.

{It is worthy noting that, in GANs~\cite{Mirza2014Conditional,Goodfellow2014Generative}, the realistic of the generated images is important. However, the purpose of DA methods is to reduce the domain discrepancy, while the realistic of the generated image is not that important. Therefore, our focus lies in the domain distribution alignment for cross-domain visual recognition instead of the pure image generation like GAN.
To this end, in this work, we proposed a simple, slim but effective \textbf{C}oupled \textbf{a}dversarial \textbf{t}ransfer based domain adaptation (CatDA) algorithm. To be specific, the proposed CatDA is formulated with a slim and symmetric multilayer perceptron (MLP) structure instead of deep structure for generative adversarial adaptation.}
CatDA comprises of two symmetric and coupled sub-networks, with each a generator, a discriminator, a domain specific term and a domain knowledge fidelity term are formulated. CatDA is then implemented by coupled learning of the two sub-networks. With the symmetry, both domains can be generated from each other with an adversarial mechanism supervised by the coupled discriminators, such that the network compatibility for arbitrary domain generation can be guaranteed.

{In order to avoid the critical flaw of high-capacity of the network mapping function during domain adversarial training, the semi-supervised mode is therefore considered in our method. In addition, a content fidelity term and a domain loss term are proposed in the generators for achieving the joint domain-knowledge preservation in source and target domains.}
The structure of CatDA can be simply described as two generators and two discriminators. Specifically, the main contribution and novelty of this work are fourfold:

\begin{itemize}
\item In order to reduce the distribution discrepancy between domains, we propose a simple but effective coupled adversarial transfer network (CatDA), which is a slim and symmetric adversarial domain adaptation network structured by shallow multilayer perceptrons (MLPs). Through the proposed network, source and target domains can be generated against each other with an adversarial mechanism supervised by the coupled discriminators.

\item Inspired by the cycleGAN, the CatDA adopts a cycling structure and formulates a generative adversarial domain adaptation framework comprising of two sub-networks with each carries a generator and a discriminator. The coupled learning algorithm follows a two-way strategy, such that arbitrary domain generation can be achieved without constraining the input to be source or target.

\item {To avoid the limitation of domain adversarial training that feature extraction function has high-capacity, in domain alignment, a semi-supervised domain knowledge fidelity loss and domain specific loss are designed for domain content self-preservation and domain realistic. In this way, domain distortion in domain generation is avoided and the domain-agnostic feature representation become more stable and discriminative.}
\item {A simple but effective MLP network is tailored to handle the small-scale cross-domain visual recognition datasets. In this way, both the requirements of large-scale samples and the pre-trained processing are avoided. Extensive experiments and comparisons on a number of benchmark datasets demonstrate the effectiveness and competitiveness of the proposed CatDA over state-of-the-art methods.}
\end{itemize}

The rest of this paper is organized as follows. In Section \Rmnum{2}, the related work in domain adaptation is reviewed. In Section \Rmnum{3}, we present the proposed CatDA method in detail. In Section \Rmnum{4}, the experiments and comparisons on a number of common datasets are presented. The discussion is presented in Section \Rmnum{5}, and finally Section \Rmnum{6} concludes this paper.

\section{Related Work}
Our approach involves domain adaptation (DA) and generative adversarial methods. Therefore, In this section, the shallow domain adaptation, deep domain adaptation, and generative adversarial networks are briefly introduced, respectively.

\subsection{Shallow Domain Adaptation}

 In recent years, a number of shallow learning approaches have been proposed in domain adaptation.
Generally, these shallow domain adaptation methods can be divided into three categories.

\textit{Classifier based approaches}. A generic way in this category is to learn a common classifier on source domain leveraging source data and a few labeled target data. In AMKL, Duan et al.~\cite{Duan2010Visual} proposed an adaptive multiple kernel learning method for video event recognition. Also, a domain transfer MKL (DTMKL)~\cite{Duan2012Domain} was proposed by jointly learning a SVM and a kernel function for classifier adaptation. Li et al.~\cite{Li2018cdelm} proposed the cross-domain extreme learning machine for visual domain adaptation~\cite{Huang2012Extreme,Lei2017Abnormal}, in which MMD was formulated for characterizing and matching the marginal and conditional distribution between domains. Zhang et al.~\cite{eda2016zhang} proposed a robust extreme domain adaptation (EDA) based classifier with manifold regularization for cross-domain visual recognition.

\textit{Feature augmentation/transformation based approaches}. In MMDT, Hoffman et al.~\cite{Hoffman2014Asymmetric} proposed a Max-Margin Domain Transforms method, in which a category specific transformation was optimized for domain transfer. Long et al.~\cite{long2013transfer} proposed a Transfer Sparse Coding (TSC) method to learn robust sparse representations, in which the empirical Maximum Mean Discrepancy (MMD)~\cite{iyer2014maximum} is constructed as the distance measure. Then he~\cite{long2014transfer} also proposed a Transfer Joint Matching approach. This TJM method aims to learn a non-linear transformation across domains by minimizing distribution discrepancy based on MMD. Zhang et al. ~\cite{Lei2016Efficient} proposed a regularized subspace alignment method for realizing cross-domain odor recognition with signal shift.

\textit{Feature reconstruction based approaches}. Different from those methods above, domain adaptation is achieved by feature reconstruction between domains. Jhuo et al.~\cite{jhuo2012robust} proposed a RDALR method, in which the source samples was reconstructed by the target domain with low-rank constraint model. Similarly, Shao et al.~\cite{shao2014generalized} proposed a LTSL method by pre-learning a subspace through principal component analysis (PCA) or linear discriminant analysis (LDA), then low-rank regularization across domains is modeled in this method. Zhang et al.~\cite{zhang2016lsdt} proposed a Latent Sparse Domain Transfer (LSDT) approach by jointly learning a common subspace and a sparse reconstruction matrix across domains, and this method achieves good results.

\begin{figure*}
\begin{center}
 \includegraphics[width=1\linewidth]{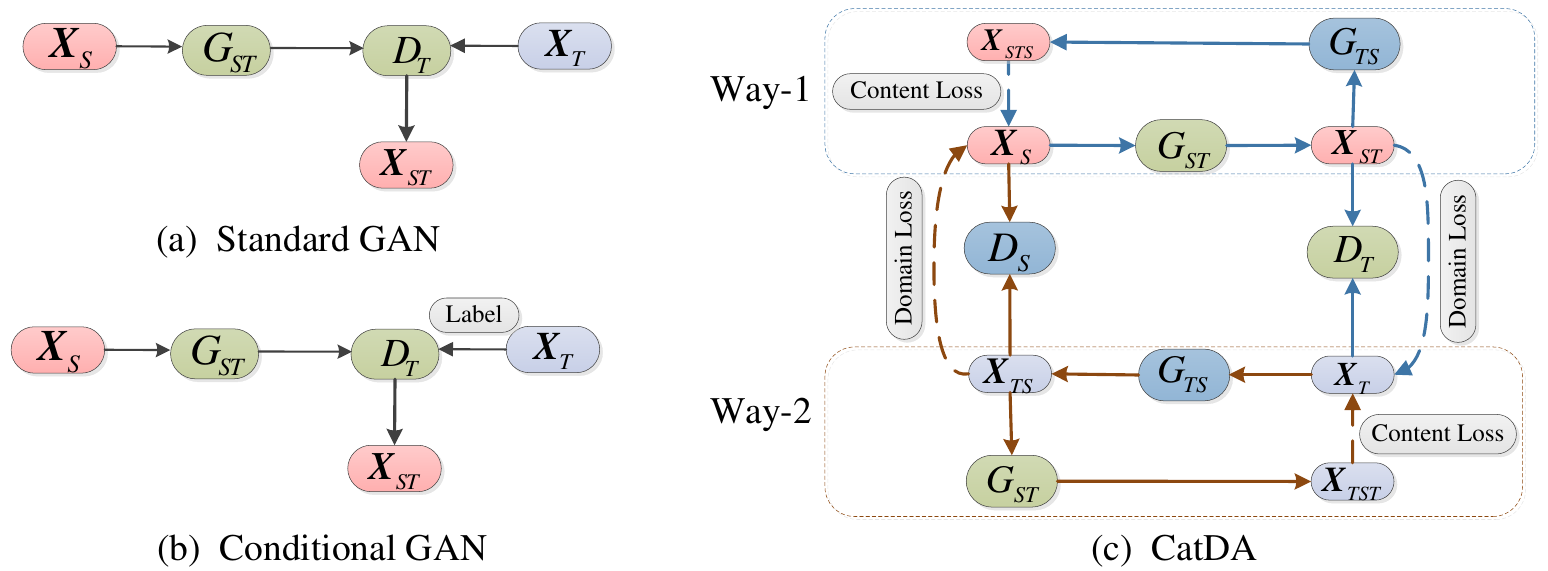}
\end{center}
   \caption{(a) The standard GAN is described. (b) The conditional GAN is structured. (c) The structure of the proposed CatDA, which is a slim but symmetric network, with each stream a generator, a discriminator, a domain loss and a content loss are contained. The two-way generation function is a bijective mapping. Besides the inherent GAN losses, the domain knowledge fidelity loss (content loss) and the domain specific loss (domain loss) are designed for domain content self-preservation and domain realistic. “$ {{\bf{X}}_S}$ ” and “$ {{\bf{X}}_T}$” represent the labeled samples in the source and target domain for training; “$ {\hat{\bf{X}}_{ST}}$ ”represents the generated data from $S\to \hat{T}$ ; “$ {\hat{\bf{X}}_{TS}}$ ”represents the generated data from $T\to \hat{S}$. We aim to learn a discriminative classifier through ${\hat{\bf{X}}_{ST}}$ as well as the corresponding labels to classify unlabeled target test data "$ {{\bf{X}}^{test}_T}$".}
%   \caption{ The structure of proposed CatDA. It is a slim but symmetric GAN, in which a generator, a discriminator, a domain loss and a content loss are contained. The two-way generation function is a bijective mapping. Besides the GAN losses, the domain knowledge fidelity loss (content loss) and the domain specific loss (domain loss) are designed for domain content self-preservation and domain realistic. “$ {{\bf{X}}_S}$ ” and “$ {{\bf{X}}_T}$” represent the labeled data in the source and target domain for training; “$ {{\bf{X}}_{ST}}$ ”represents the generated data from $S\to T$ ;“$ {{\bf{X}}_{TS}}$ ”represents the generated data from $T\to S$. We aim to learn a discriminative classifier through ${{\bf{X}}_{ST}}$ as well as the corresponding labels to classify unlabeled target test data "$ {{\bf{X}}^{test}_T}$".}
\label{fig1}
\end{figure*}
\subsection{Deep Domain Adaptation}
The other category of data-driven domain adaptation method is deep learning and it has witnessed a series of great success~ \cite{tzeng2015simultaneous,oquab2014learning,xie2015transfer,Ma2018Object,Lu2018DNN}. However, for small-data tasks, deep learning method may not improve the performance too much. Thus, recently, deep domain adaptation methods under small-scale tasks have been emerged.

Donahue et al.~\cite{donahue2014decaf} proposed a deep transfer strategy leveraging a CNN network  on the large-scale ImageNet for small-scale object recognition tasks. Similarly, Razavian et al.~\cite{sharif2014cnn} also proposed to train a deep network based on ImageNet for high-level domain feature extractor.
Tzeng et al.~\cite{tzeng2015simultaneous} proposed a CNN network based on DDC method both aligning domains and tasks. In DAN, Long et al.~\cite{long2015learning} proposed a deep transfer network leveraging MMD loss on the high-level features between the two-streamed fully-connected layers from two domains. Then he also proposed another two famous methods. In RTN~\cite{long2016unsupervised}, he proposed a residual transfer network which aims to learn a residual function between the source and target domain. A joint adaptation networks (JAN) is another method which was proposed~\cite{Long2016Deep} to learn a adaptation network. This network aligned the joint distributions across domains with a joint maximum mean discrepancy (JMMD) criterion.
%Oquab et al.~\cite{oquab2014learning} proposed a CNN architecture for middle-level feature transfer, which was trained on a large-scale annotated image set.
Hu et al.~\cite{hu2015deep} proposed a deep transfer metric learning (DTML) method leveraging the MLPs instead of CNN.
The novelty of this method is to learn a set of hierarchical nonlinear transformations. Autoencoder is an unsupervised feature representation~\cite{Yang2016Autoencoder} and Wen et al.~\cite{Long2018A} proposed a deep autoencoder based feature reconstruction for domain adaptation, which aims to share the feature representation parameters between source and target domains. Recently, Chen et al.~\cite{Chen2018Broad} proposed a broad learning system instead of deep system which can also be considered for transfer learning.

\subsection{Generative Adversarial Networks}

The generative adversarial network (GAN) was first proposed by Goodfellow et al.~\cite{Goodfellow2014Generative} to generate images and produced a high impact in deep learning. GAN generally comprises of two operators: a generator (G) and a discriminator (D). The discriminator discriminates whether the sample is fake or real, while the generator produces samples as real as possible to cheat the discriminator. Mirza et al.~\cite{Mirza2014Conditional} proposed a conditional generative adversarial net (CGAN) where both networks G and D receive an additional information vector as input. Salimans et al.~\cite{Salimans2016Improved} achieved state-of-the-art results in semi-supervised classification and improves the visual realistic and image quality compared to GAN. Zhu et al.~\cite{Zhu2017Unpaired} proposed a cycleGAN for discovering cross-domain relations and transferring style from one domain to another, which was similar with DiscoGAN~\cite{Kim2017Learning} and DualGAN~\cite{Yi2017DualGAN}. The key attributes such as orientation and face identity are preserved.

In~\cite{ganin2017domain}, DANN is one of the first work in deep domain adaptation, in which the adversarial mechanism of GAN was used to bridge the gap between the source and target domains. A novel ADDA method is proposed in ~\cite{Tzeng2017Adversarial} for adversarial domain adaptation. In this method, the convolutional neural network (CNN) was used for adversarial discriminative feature learning. A GAN based model~\cite{Bousmalis2016Unsupervised} that adapted the source domain images to appear to be drawn from the target domain was proposed, in which domain image generation was focused. The three works have shown the potential of adversarial learning in domain adaptation. Additionally, a CyCADA method~\cite{judy2017CYCADA} was proposed for cycle generation, in which the representations in both pixel-level and feature-level are adaptive by enforcing cycle-consistency.
\section{The Proposed CatDA}
\subsection{Notation}
In our method, the source domain is defined by subscript ``$S$'' and target domain is defined by subscript ``$T$'', respectively. The training set of source and target domain are defined as $ {{\bf{X}}_S} $  and $ {{\bf{X}}_T} $, respectively.  $\bm{Y}$ denotes the data labels. A generator network is denoted as $G_{ST}$: $S \to \hat{T}$, that embeds data from source domain $S$ to its co-domain $\hat{T}$. The discriminator network is denoted as $D_{T}$, which tries to discriminate the real samples in target domain and the generated samples in co-domain $\hat{T}$. Similarly, $G_{TS}:$ $T \to \hat{S}$ aims to map the data from target domain $T$ to its co-domain $\hat{S}$, and $D_{S}$ tries to discriminate the real samples in source domain and the generated samples in co-domain $\hat{S}$.
 “${\hat{\bf{X}}}_{ST}$ ”represents the generated target data from $S\to \hat{T}$ ; “$ {\hat{\bf{X}}_{TS}}$ ”represents the generated source data from $T\to \hat{S}$. $ {{\bf{X}}^{test}_T}$ represents the unlabeled target test data.

\subsection{Motivation}
Direct supervised learning of an effective classifier on the target domain is not allowed due to the label scarcity. {Therefore, in this paper, we would like to answer whether the target data can be generated by using the labeled source data, such that the classifier can be trained on the generated target data with labels. Our key idea is to learn a "source $\to$ target" generative feature representation $ {\hat{\bf{X}}}_{ST}$ through an adversarial domain generator, then a domain-agnostic classifier can be learnt on the generated features for cross-domain applications. Noteworthily, our aim is to minimize the feature discrepancy between domains via similarly distributed feature generation rather than generating a vivid target image. Therefore, a simple and flexible network is much more expected for homogeneous image feature generation, instead of very complicated structure (encoder vs. decoder) for realistic image rendering.}

Visually, the structure of the standard GAN and conditional GAN are shown in Fig.~\ref{fig1} (a) and  Fig.~\ref{fig1} (b), respectively. There are several limitations for the two models. In standard GAN, explicitly supervised data is seldom available, and the randomly generated samples can become tricky if the content information is lost. Thus the trained classifier may not work obviously well. In conditional GAN, although a label constraint is imposed, it does not guarantee the cross-domain relation because of the one-directional domain mapping.
Since conditional GAN architecture only learns one mapping from one domain to another (one-way), a two-way coupled adversarial domain generation method with more freedom is therefore proposed in this paper, as shown in Fig.~\ref{fig1} (c). The core of our CatDA model depends on two symmetric GANs, which then result in a pair of symmetric generative and discriminative functions. The two-way generator function can be recognized as a bijective mapping, and the work flow of the proposed CatDA in implementation can be described as follows.

First, different from GAN, the image or feature instead of noise is fed as input into the model. The way-1 of CatDA comprises of the generator $G_{ST}$ and the discriminator $D_{T}$. The way-2 comprises of the generator $G_{TS}$ and the discriminator $D_{S}$. For way-1, the source data ${{\bf{X}}_S}$ is fed into the generator, and the co-target data $ {\hat{\bf{X}}}_{ST}$ is generated. Then the generated target data $ {\hat{\bf{X}}}_{ST}$ and the real target data ${{\bf{X}}_T}$ are fed into the discriminator network $D_{T}$ for adversarial training. For way-2, the similar operation with way-1 is conducted. In order to achieve the bijective mapping, we expect that the real source data can be recovered by feeding the generated ${\hat{\bf{X}}}_{ST}$ into the generator $G_{TS}$ for progressively learning supervised by $D_{S}$. Similarly, $G_{ST}$ is also fine-tuned by feeding the ${\hat{\bf{X}}}_{TS}$ supervised by $D_{T}$ to recover the real target training data.

\subsection{Model Formulation}
\begin{figure}[t]
\begin{center}
 \includegraphics[width=1\linewidth]{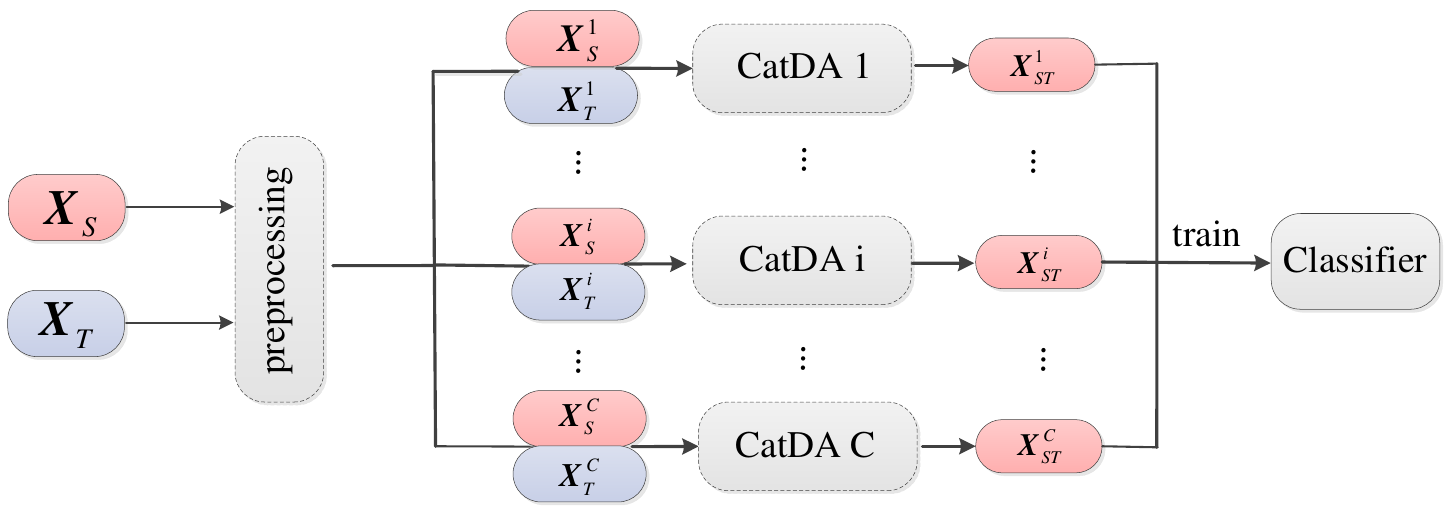}
\end{center}
   \caption{The pipeline of the class-wise CatDA. In order to avoid the domain adversarial training limitation that feature extraction function has high-capacity, the semi-supervised method is used in our proposed CatDA to preserve the raw content information in the generated samples. We process the samples class by class which results in a semi-supervised structure. The training samples in source domain and target domain per class are preprocessed independently. Note that “${i}$” represents the index of class in source and target and “${C}$” denotes the number of classes. The number of networks to be trained equals to number $C$ of classes.}
\label{pipeline}
\end{figure}
\begin{figure}[t]
\begin{center}
 \includegraphics[width=1\linewidth]{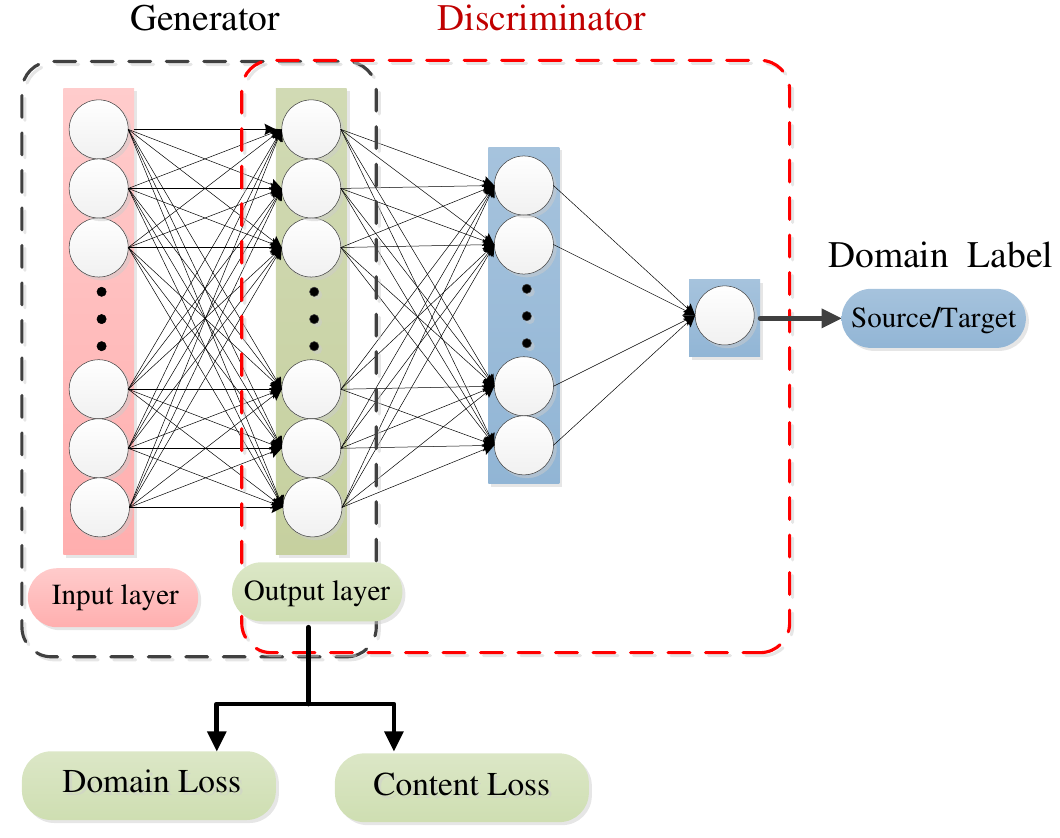}
\end{center}
   \caption{The network structure of the adversarial training model in CatDA. The input sample for each class is fed into the generator (two-layered perceptrons), and then the generated sample is sent to the discriminator (three-layered perceptrons) for classifying whether it belongs to the target domain. In order to reduce the distribution mismatch across the generated samples and the real samples, a domain specific loss (i.e. domain loss) is imposed. Further, for preserving the content information of each sample, a domain knowledge fidelity term (i.e. content loss) is established in our model.}
\label{fig2}
\end{figure}

In order to avoid the domain adversarial training limitation of high-capacity of the feature extraction function, the semi-supervised strategy is used in our proposed CatDA to preserve the content information of the generated samples. We process the samples class by class which results in a semi-supervised structure. The training data in source domain and target domain per class are preprocessed independently, thus the number of networks to be trained equals to the number of classes. Specifically, the pipeline of the class-wise CatDA is shown in Fig.~\ref{pipeline}, and the class conditional information is used.
For each CatDA model, the generator is a two-layered perceptron and the discriminator is a three-layered perceptron. Sigmoid function is selected as the activation function in hidden layers. The network structure of the joint generator and discriminator is described in Fig.~\ref{fig2}.

The proposed CatDA model has a symmetric structure comprising of two generators and two discriminators, which are described as two ways across domains ($S \to \hat{T}$ and $T \to \hat{S}$). We first describe the model of way-1 ($S \to \hat{T}$), which shares the same model and algorithm as way-2 ($T \to \hat{S}$).
\begin{itemize}
\item Way-1: $S \to \hat{T}$:
\end{itemize}

A target domain discriminator $D_{T}$ is formulated to classify whether a generated data is drawn from the target domain (real). Thus, the discriminator loss ${L_{D_T}}$ is formulated as
\begin{equation}
\begin{split}
\min_{D_T}{L_{D_T}} =  &- {E_{{\bm{X}_T} \sim {{\bm{\mathcal{X}}}_T}}}[log{{{D}}_T}({{\bf{X}}_T})]\\
 &- {E_{{\bm{X}_S} \sim {{\bm{\mathcal{X}}}_S}}}[log(1 - {{{D}}_T}({\hat{\bf{X}}}_{ST}))]\\
\end{split}
\label{eequa1}
\end{equation}
where ${\hat{\bf{X}}}_{ST} = {G_{ST}}({{\bf{X}}_S})$. $G_{ST}$ is the generator for generating realistic data similar to target domain. Therefore, the supervised generator loss ${L_{G_{ST}}}$ is formulated as
\begin{equation}
\begin{split}
\min_{G_{ST}}{L_{G_{ST}}} = & - {E_{{\bm{X}_S} \sim{{\bm{\mathcal{X}}}_S}}}[log{{{D}}_T}({\hat{\bf{X}}}_{ST})]\\
 = & - {E_{{\bm{X}_S} \sim{{\bm{\mathcal{X}}}_S}}}[log{{{D}}_T}({G_{ST}}({{\bf{X}}_S})]\\
\end{split}
\label{eequa2}
\end{equation}

The two losses in Eq.~(\ref{eequa1}) and (\ref{eequa2}) are the inherent loss functions in traditional GAN models. The focus of CatDA is to reduce the distribution difference across domains. Therefore, in the proposed CatDA model, two novel loss functions are proposed, which are the domain specific loss (domain loss) and the domain knowledge fidelity loss (content loss).

One of the feasible strategies for reducing the domain discrepancy is to find an abstract feature representation under which the source and target domains are similar. This idea has been explored in \cite{long2015learning,long2016unsupervised,Long2016Deep} by leveraging the Maximum Mean Discrepancy (MMD) criterion, which is used when the source and target data distributions differs. Therefore, in this paper, we focus on the domain specific loss by a simple and approximate MMD, which is formalized to maximize the two-sample test power and minimize the Type II error (the failure of rejecting a false null hypothesis). For convenience, we define the proposed domain loss as ${L_{DomainT}}$, which is minimized to help the learning of the generator $G_{ST}$ as shown in Fig. \ref{fig1} (c). Specifically, in order to reduce the distribution mismatch between the generated target data ${\hat{\bf{X}}}_{ST}$ and the original target data ${\bf{X}}_T$, the domain specific loss (domain loss) can be formulated as
\begin{equation}
\begin{split}
\min_{G_{ST}}{L_{DomainT}} =& {\left\| {{\overline {\hat{\bf{X}}}_{ST}} - {{\overline {\bf{X}} }_T}} \right\|^2}\\
\end{split}
\label{eequa3}
\end{equation}
where $||\cdot||$ denotes the $l_2$-norm, ${\overline {\hat{\bf{X}}}_{ST}}$ and ${{\overline {\bf{X}} }_T}$ represents the center of the co-target data and target data, respectively. Noteworthily, during the network training phase, the sigmoid function is imposed on the domain loss for probability output normalized to $[0,1]$. Therefore, the target domain loss shown in Eq.~(\ref{eequa3}) can be further written as
\begin{equation}
\begin{split}
\mathop {\min }\limits_{{G_{ST}}}& {L_{DomainT}} = 1/(1 + exp( - {{\left\| {{\overline{\hat{\bf{X}}}_{ST}} - {{\overline {\bf{X}} }_T}} \right\|}^2}))\\
%&s.t.\left\{ {\begin{array}{*{20}{c}}
%{a = {{\left\| {{{\bf{X}}_{ST}} - {{\overline {\bf{X}} }_T}} \right\|}^2}}\\
%{{{\bf{X}}_{ST}} = {G_{ST}}({{\bf{X}}_S})}
%\end{array}} \right.
\end{split}
\label{eequa4}
\end{equation}

\begin{table}%\footnotesize
\begin{tabular}{l}
\toprule
\textbf{Algorithm 1}  The Proposed CatDA\\
\midrule
\textbf{Input:} \hspace*{0.1cm} Source data ${{\bf{X}}_S}=[{{\bf{X}}^{1}_S},...{{\bf{X}}^{C}_S}] $, \\
\hspace*{1cm} target training data ${{\bf{X}}_T}=[{{\bf{X}}^{1}_T},...{{\bf{X}}^{C}_T}] $, \\
%\hspace*{0.9cm} source label $\bm{Y}_S$, target label $\bm{Y}_T$,\\
\hspace*{1cm} iterations $maxiter1$, $maxiter2$, the number “$ {C}$ ” of classes;\\
\textbf{Procedure:}\\
%1. Preprocess to get ${{\bf{X}}^{i}_S}$, ${{\bf{X}}^{i}_T}$,\\
%\hspace*{0.9cm} where the “$ {i}$ ” represents the $ {i}^{th}$ class of samples, \\
%\hspace*{0.9cm} the “$ {C}$ ” means the number of classes.\\
\textbf{For} $i=1,...C$ \textbf{do}  \\
1.\hspace*{0.1cm}  \textbf{{Initialize ${\hat{\bf{X}}^{i}_{ST}}$ and ${\hat{\bf{X}}^{i}_{TS}}$ using traditional GAN}}:\\
\hspace*{0.2cm}  \textbf{While} $iter1<maxiter1$ \textbf{do} \\
\hspace*{0.4cm}	\textbf{Step1}: Train the generator $G_{ST}$ and discriminator $D_{T}$ by solving \\
\hspace*{1cm}   problem (\ref{eequa1}) and (\ref{eequa2}) using back-propagation (BP) algorithm;\\
\hspace*{0.4cm}	\textbf{Step2}: Compute ${\hat{\bf{X}}}^{i}_{ST} = {G_{ST}}({{\bf{X}}^{i}_S})$;\\

\hspace*{0.4cm}	\textbf{Step3}: Train the generator $G_{TS}$ and discriminator $D_{S}$ by solving\\
\hspace*{1cm}   the problem (\ref{eequa7}) using BP algorithm; \\
\hspace*{0.4cm}	\textbf{Step4}: Compute ${\hat{\bf{X}}}^{i}_{TS} = {G_{TS}}({{\bf{X}}^{i}_T})$;   \\
\hspace*{0.4cm} $iter1=iter1+1$;\\
%\hspace*{0.4cm}	Check Convergence.\\
\hspace*{0.2cm} \textbf{end while}\\

2.\hspace*{0.1cm}  \textbf{{Update  ${\hat{\bf{X}}^{i}_{ST}}$ and ${\hat{\bf{X}}^{i}_{TS}}$ using the proposed model}}:\\
 \hspace*{0.2cm} \textbf{While} $iter2<maxiter2$ \textbf{do} \\
\hspace*{0.4cm}	\textbf{Step1}: Train the generator $G_{ST}$ and the discriminator $D_{T}$, by \\
\hspace*{1cm}  solving the problem (\ref{eequa1}) and (\ref{eequa6}) using BP algorithm; \\

\hspace*{0.4cm}	\textbf{Step2}: Update ${\hat{\bf{X}}^{i}_{ST}}={G_{ST}}({{\bf{X}}^{i}_S})$ and ${\hat{\bf{X}}^{i}_{TST}}={G_{ST}({\hat{\bf{X}}}^{i}_{TS}))}$; \\

\hspace*{0.4cm}	\textbf{Step3}: Train the generator $G_{TS}$ and the discriminator $D_{S}$, by    \\
\hspace*{1cm} solving the problem (\ref{eequa7}) and (\ref{eequa8}) using BP algorithm;\\

\hspace*{0.4cm}	\textbf{Step4}: Update ${\hat{\bf{X}}^{i}_{TS}}={G_{TS}}({{\bf{X}}^{i}_T})$ and ${\hat{\bf{X}}^{i}_{STS}}={G_{TS}}({\hat{\bf{X}}}^{i}_{ST})$; \\

\hspace*{0.4cm} $iter2=iter2+1$;\\
%\hspace*{0.4cm}	Check Convergence.\\
\hspace*{0.2cm}\textbf{end while}\\
\textbf{End}\\
\textbf{Output:} ${\hat{\bf{X}}_{ST}}$.\\
\bottomrule
\end{tabular}
\end{table}

Additionally, for preserving the content in source data, we establish a content fidelity term in our model. Ideally, the equality ${G_{TS}}({G_{ST}}({{\bf{X}}_S})) = {{\bf{X}}_S}$ should be satisfied, that is, the generation is reversible. However, this hard constraint is difficult to be guaranteed and thus a relaxed soft constraint is more desirable. To this end, we try to minimize the $l_2$ distance between ${G_{TS}}({G_{ST}}({{\bf{X}}_S}))$ and ${{\bf{X}}_S}$ and formulate a content loss function ${L_{CON1}}$, i.e. source content loss as follows
\begin{equation}
\begin{split}
\min_{G_{TS}}&{L_{CON1}} = 1/(1 + exp( - {{\left\| {{\hat{\bf{X}}_{STS}} - {{\bf{X}}_S}} \right\|}^2}))\\
%&s.t.\left\{ {\begin{array}{*{20}{c}}
%{b = {{\left\| {{{\bf{X}}_{STS}} - {{\bf{X}}_S}} \right\|}^2}}\\
%{{{\bf{X}}_{STS}}{\rm{ = }}{G_{TS}}({G_{ST}}({{\bf{X}}_S}))}
%\end{array}} \right.
\end{split}
\label{eequa5}
\end{equation}
where ${{\hat{\bf{X}}_{STS}}{\rm{ = }}{G_{TS}}({G_{ST}}({{\bf{X}}_S}))}$, and $G_{TS}$ is a generator of way-2 ($T \to \hat{S}$).
Finally, the objective function of the way-1 generator is composed of 3 parts:
\begin{equation}\small
\begin{split}
\mathop {\min_{G_{ST},G_{TS}}} &{L_{Way1}}
= {L_{G_{ST}}} + {L_{DomainT}} + {L_{CON1}}
\end{split}
\label{eequa6}
\end{equation}

\begin{itemize}
\item Way-2: $T \to \hat{S}$:
\end{itemize}

For way-2, the similar models with way-1 are formulated, including the source discriminator loss $L_{D_S}$, the source data generator loss $L_{G_{TS}}$, source domain loss $L_{DomainS}$, and the target content loss $L_{CON2}$. Specifically, the loss functions of way-2 can be formulated as follows
\begin{equation}\small
%\begin{eqnarray}
\begin{split}
&\mathop {\min }\limits_{{D_S}}{L_{{D_S}}}= - {E_{{{\bf{X}}_S} \sim {{\bm{\mathcal{X}}}_S}}}[log{{{D}}_S}({{\bf{X}}_S})]\\
&\hspace*{1.8cm}- {E_{{{\bf{X}}_T} \sim {{\bm{\mathcal{X}}}_T}}}[log(1 - {{{D}}_S}({\hat{\bf{X}}_{TS}}))]\\
&\mathop {\min }\limits_{{G_{TS}}} {L_{G_{TS}}} =  - {E_{{{\bf{X}}_T} \sim {{\bm{\mathcal{X}}}_T}}}[log{{{D}}_S}({\hat{\bf{X}}_{TS}})]\\
&\mathop {\min }\limits_{{G_{TS}}} {L_{DomainS}} = 1/(1 + exp( - {{\left\| {{\overline{\hat{\bf{X}}}_{TS}} - {{\overline {\bf{X}} }_S}} \right\|}^2}))\\
&\mathop {\min }\limits_{{G_{ST}}} {L_{CON2}} = 1/(1 + exp( - {{\left\| {{\hat{\bf{X}}_{TST}} - {{\bf{X}}_T}} \right\|}^2}))\\
%s.t.\left\{ {\begin{array}{*{20}{c}}
%{a = {{\left\| {{{\bf{X}}_{TS}} - {{\overline {\bf{X}} }_S}} \right\|}^2}}\\
%{{{\bf{X}}_{TS}} = {G_{TS}}({{\bf{X}}_T})}\\
%{b = {{\left\| {{{\bf{X}}_{TST}} - {{\bf{X}}_T}} \right\|}^2}}\\
%{{{\bf{X}}_{TST}}{\rm{ = }}{G_{ST}}({G_{TS}}({{\bf{X}}_T}))}\\
%\end{array}} \right. \\
\end{split}
\label{eequa7}
%\end{eqnarray}
\end{equation}
%\begin{eqnarray*}
%s.t.\left\{ {\begin{array}{*{20}{c}}
%{a = {{\left\| {{{\bf{X}}_{TS}} - {{\overline {\bf{X}} }_S}} \right\|}^2}}\\
%{{{\bf{X}}_{TS}} = {G_{TS}}({{\bf{X}}_T})}\\
%{b = {{\left\| {{{\bf{X}}_{TST}} - {{\bf{X}}_T}} \right\|}^2}}\\
%{{{\bf{X}}_{TST}}{\rm{ = }}{G_{ST}}({G_{TS}}({{\bf{X}}_T}))}\\
%\end{array}} \right. \\
%\end{eqnarray*}
where ${{\hat{\bf{X}}_{TS}} = {G_{TS}}({{\bf{X}}_T})}$ and ${{\hat{\bf{X}}_{TST}}{\rm{ = }}{G_{ST}}({G_{TS}}({{\bf{X}}_T}))}$. ${\overline {\hat{\bf{X}}}_{TS}}$ and ${{\overline {\bf{X}} }_S}$ are the centers of the generated source data and the real source data.
Similar to Eq.~(\ref{eequa6}), the objective function of the way-2 generator can be formulated as
%follows
\begin{equation}\small
\begin{split}
\min_{G_{TS},G_{ST}} &{L_{Way2}}
= {L_{G_{TS}}} + {L_{DomainS}} + {L_{CON2}}
\end{split}
\label{eequa8}
\end{equation}

\begin{itemize}
\item Complete CatDA Model:
\end{itemize}

The proposed CatDA model is a coupled net of the way-1 and way-2, each of which learns the bijective mapping from one domain to another. The two ways in CatDA are jointly trained in an alternative manner. The generated data ${\hat{\bf{X}}_{ST}}$ and ${\hat{\bf{X}}_{TS}}$ are fed into the discriminators $D_{T}$ and $D_{S}$, respectively.

By joint learning of the Way-1 and Way-2, the complete model of CatDA including the generator $G$ and the discriminator $D$ can be formulated as the follows.
\begin{equation}\small
\begin{split}
\min_{G_{ST},G_{TS}}{L_G} = &{L_{Way1}} + {L_{Way2}}\\
% = &{L_{GenT}} + {L_{DomainT}} + {L_{STS}} + {L_{GenS}} + {L_{DomainS}} + {L_{TST}}\\
 \min_{D_{T},D_{S}}{L_D} =& {L_{{D_T}}} + {L_{{D_S}}}
\end{split}
\label{eequa9}
\end{equation}

In detail, the complete training procedure of the proposed CatDA approach is summarized in~\textbf{Algorithm 1}.

\subsection{Classification}
For classification, the general classifiers can be trained by the domain aligned and augmented training samples $[\hat{\bm{X}}_{ST},\bm{X}_T]$ with label $\bm Y=[\bm{Y}_S,\bm{Y}_T]$. Note that $\hat{\bm{X}}_{ST}$ is the output of \textbf{Algorithm 2}. Finally, the recognition accuracy of the unlabeled target test data ${{\bf{X}}^{test}_T}$ is reported and compared.%$\bm{X}^{test}_T$
%(e.g., SVM, least square method~\cite{kanamori2009least}, SRC~ \cite{wright2009robust})

The whole procedure of proposed CatDA for cross-domain visual recognition is clearly described in \textbf{Algorithm 2}, following which the experiments are then conducted to verify the effectiveness and superiority of the proposed method.

\begin{table}%\footnotesize
\begin{tabular}{l}
\toprule
\textbf{Algorithm 2}  The Proposed Cross-domain Visual Recognition Method\\
\midrule
\textbf{Input:} Source data ${{\bf{X}}_S} $, a very few target training data $ {{\bf{X}}_T}$,\\
 \hspace*{0.9cm} source label $\bm{Y}_S$, target label $\bm{Y}_T$, target test data $ {{\bf{X}}^{test}_T}$;\\
\textbf{Procedure:}\\
1. \textbf{Step1}: Compute ${\hat{\bf{X}}_{ST}}$ by CatDA method using \textbf{Algorithm 1};\\
2. \textbf{Step2}: Train the classifier $f(\cdot)$ on augmented training data \\
 \hspace*{0.9cm} $[\hat{\bm{X}}_{ST},\bm{X}_T]$ with label $\bm Y=[\bm{Y}_S,\bm{Y}_T]$; \\
3. \textbf{Step3}: Predict the label $\hat{\bf{Y}}$ by the classifier, i.e. $\hat{\bf{Y}}=f({{\bf{X}}^{test}_T})$. \\
\textbf{Output:} Predicted label $\hat{\bf{Y}}$.\\
\bottomrule
\end{tabular}
\end{table}

\subsection{Remarks}
{In CatDA, one key difference from the previous GAN model is that a two-way coupled architecture is proposed, with each a domain loss and a content loss are designed for domain alignment and content fidelity. Note that, the proposed CatDA has a similar structure with the cycleGAN, but essentially different in the following aspects.}
\begin{itemize}
\item {The purpose of CatDA aims at achieving domain adaptation in feature representation for cross-domain application by domain alignment and content preservation, rather than generating realistic pictures. Therefore, a simple yet effective shallow multilayer perceptrons model instead of deep model is proposed in our approach.}
\item {In order to avoid the limitation of domain adversarial training that feature extraction function has high-capacity, the semi-supervised domain adaptation model is adopted in this paper to help preserve the rich content information.}
\item {For minimizing the domain discrepancy but preserve the content information, a novel domain loss and a content loss are designed. The cycleGAN focus on image generation but not for cross-domain visual recognition.}
\end{itemize}

\section{Experiments}
\subsection{Experimental Protocol}
In our method, the total number of layers is set as 3. The neurons number of the output layer in the generator network is the same as the number of input neurons (i.e. feature dimensionality). Then, the output is fed into the discriminator network where the neurons number of hidden layer is set as 100. The neurons number in the output layer of the discriminator is set as 1. The network weights can be optimized by gradient descent based back-propagation algorithm.

\subsection{Compared Methods}
The proposed model is flexible and simple, which is therefore regarded as a shallow domain adaptation approach. Therefore, both the shallow features (e.g., pixel-level~\cite{Hoffman2016FCNs}, hand-crafted feature) and deep features can be fed as input.

In the shallow protocol, the following typical DA approaches are exploited for performance comparison.
\begin{itemize}
\item {No Adaptation (NA): a naive baseline that learns a linear SVM classifier on the source domain and then applies it to the target domain, which could be regarded as a lower bound};

\item Geodesic Flow Kernel (GFK)~\cite{Gong2012Geodesic}: a classic cross-domain subspace learning approach that aligns two domains with a geodesic flow in a Grassmann manifold;

\item Subspace Alignment (SA)~\cite{Fernando2014Unsupervised}: a widely-used feature alignment approach that projects the source subspace to the target subspace computed by PCA;

\item Robust Domain Adaptation via Low Rank (RDALR)~\cite{Chang2013Robust}: a transfer approach by reconstructing the rotated source data with the target data via low-rank representation;
\item Low-Rank Transfer Subspace Learning (LTSL)~\cite{shao2014generalized}: a subspace-based reconstruction approach that tends to align the source subspace and the target by using low-rank representation, which is similar with RADLR~\cite{Chang2013Robust};
\item Discriminative Transfer (DTSL)~\cite{Xu2015}: a subspace-based reconstruction approach that tends to align the source to the target by joint low rank and sparse representation;
\item Latent Sparse Domain Transfer (LSDT)~\cite{zhang2016lsdt}: a reconstruction-based latent sparse domain transfer method that tends to jointly learning the subspace and the sparse representation.
\end{itemize}

{{In Office-31 recognition, {{we followed the ADGANet~\cite{Sankaranarayanan2017Generate} to compare with some recent deep DA approaches,such as
DDC~\cite{tzeng2014deep}, DAN~\cite{long2015learning}, RTN~\cite{long2016unsupervised}, DANN~\cite{ganin2017domain}, ADDA~\cite{Tzeng2017Adversarial}, JAN~\cite{long2017deep} and ADGANet~\cite{Sankaranarayanan2017Generate}. For fair comparison, the deep features of Office-31 dataset extracted from ResNet-50 with convolutional neural network architecture.  Additionally, we also compared with some shallow semi-supervised methods following the~\cite{GopalanUnsupervised}.

In the deep protocol for cross-domain handwritten digits recognition, we have compared with the recent deep DA approaches, such as DANN~\cite{ganin2017domain}, Domain confusion~\cite{tzeng2015simultaneous}, CoGAN~\cite{Liu2016Coupled}, ADDA~\cite{Tzeng2017Adversarial}and ADGANet~\cite{Sankaranarayanan2017Generate}}}. Note that for fair comparison, the deep features of handwritten digit datasets extracted using LeNet with convolutional neural network architecture are fed as the input of our CatDA model.}}

\subsection{Comparison with Shallow Domain Adaptation}
\begin{figure}
  \begin{minipage}[t]{0.5\linewidth} % 如果一行放2个图，用0.5，如果3个图，用0.33
   \centering
    \includegraphics[width=0.3\linewidth,height=1.3cm]{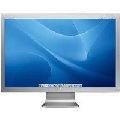}
     \includegraphics[width=0.3\linewidth,height=1.3cm]{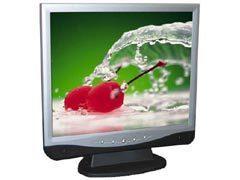}
      \includegraphics[width=0.3\linewidth,height=1.3cm]{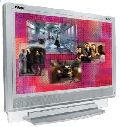}
       \includegraphics[width=0.3\linewidth,height=1.3cm]{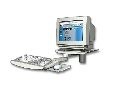}
        \includegraphics[width=0.3\linewidth,height=1.3cm]{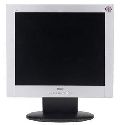}
         \includegraphics[width=0.3\linewidth,height=1.3cm]{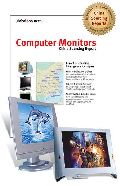}
  \caption*{ Caltech 256 }
   \label{fig:side:a}
  \end{minipage}
  \begin{minipage}[t]{0.49\linewidth}
   \centering
    \includegraphics[width=0.3\linewidth,height=1.3cm]{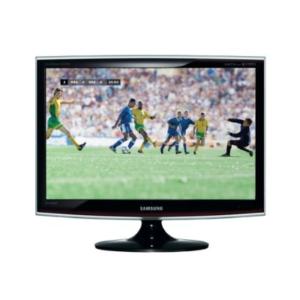}
     \includegraphics[width=0.3\linewidth,height=1.3cm]{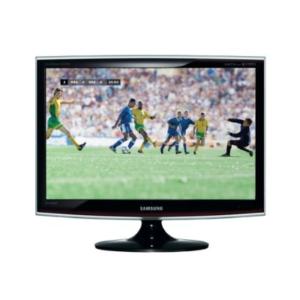}
      \includegraphics[width=0.3\linewidth,height=1.3cm]{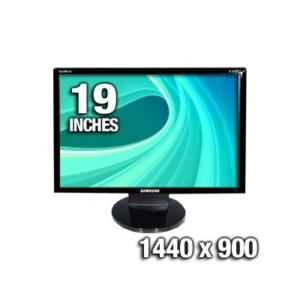}
       \includegraphics[width=0.3\linewidth,height=1.3cm]{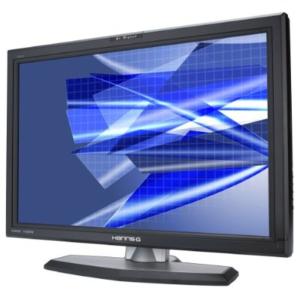}
        \includegraphics[width=0.3\linewidth,height=1.3cm]{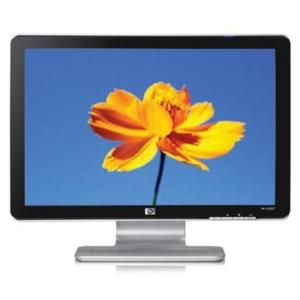}
         \includegraphics[width=0.3\linewidth,height=1.3cm]{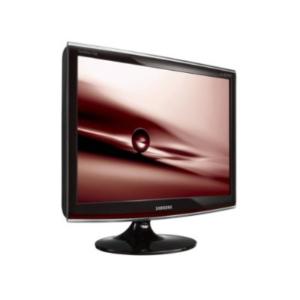}
    \caption*{Amazon}
    \label{fig:side:b}
  \end{minipage}
    \begin{minipage}[t]{0.49\linewidth}
   \centering
    \includegraphics[width=0.3\linewidth,height=1.3cm]{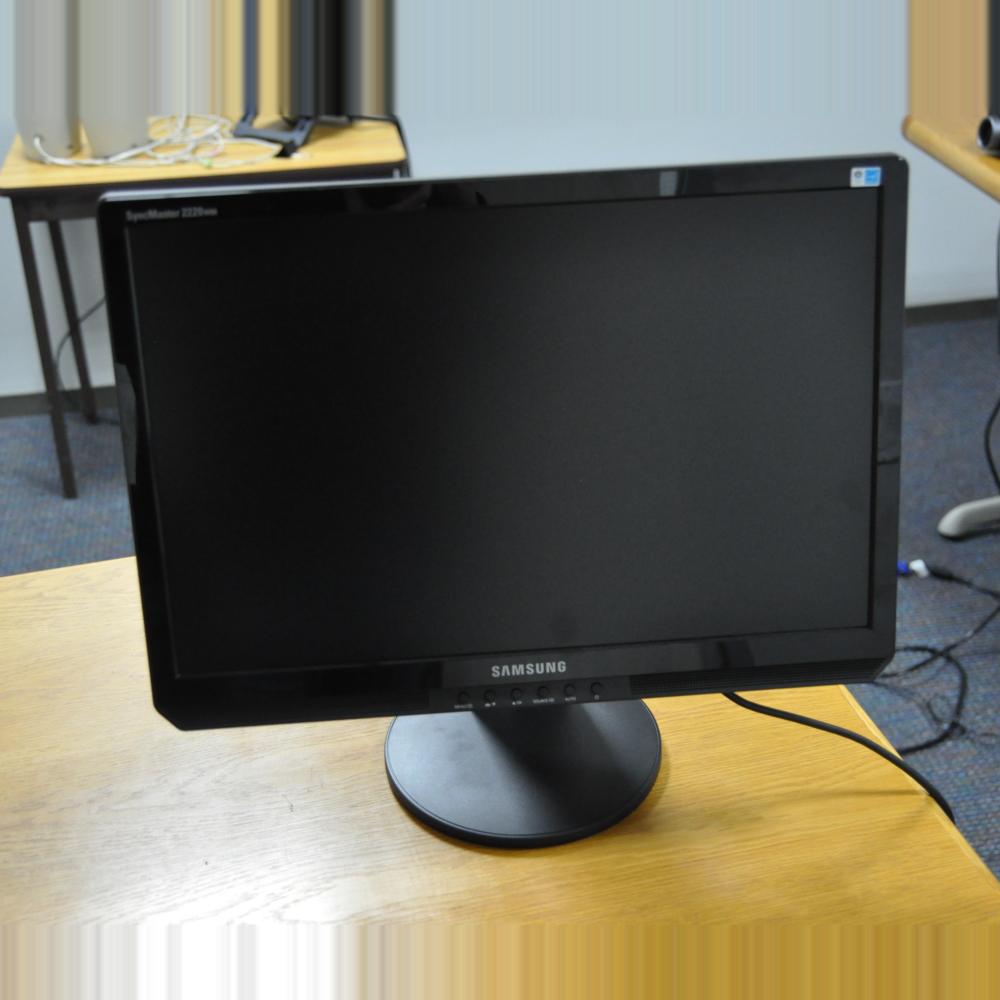}
     \includegraphics[width=0.3\linewidth,height=1.3cm]{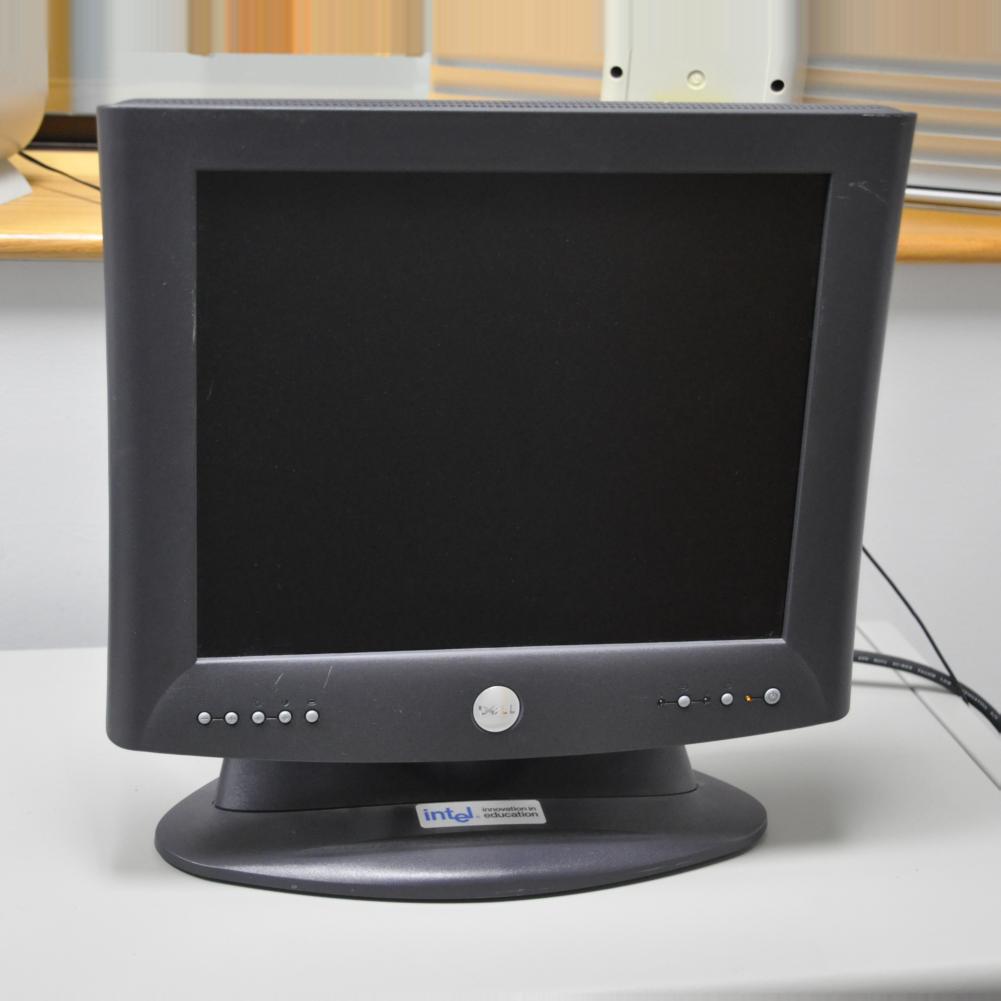}
      \includegraphics[width=0.3\linewidth,height=1.3cm]{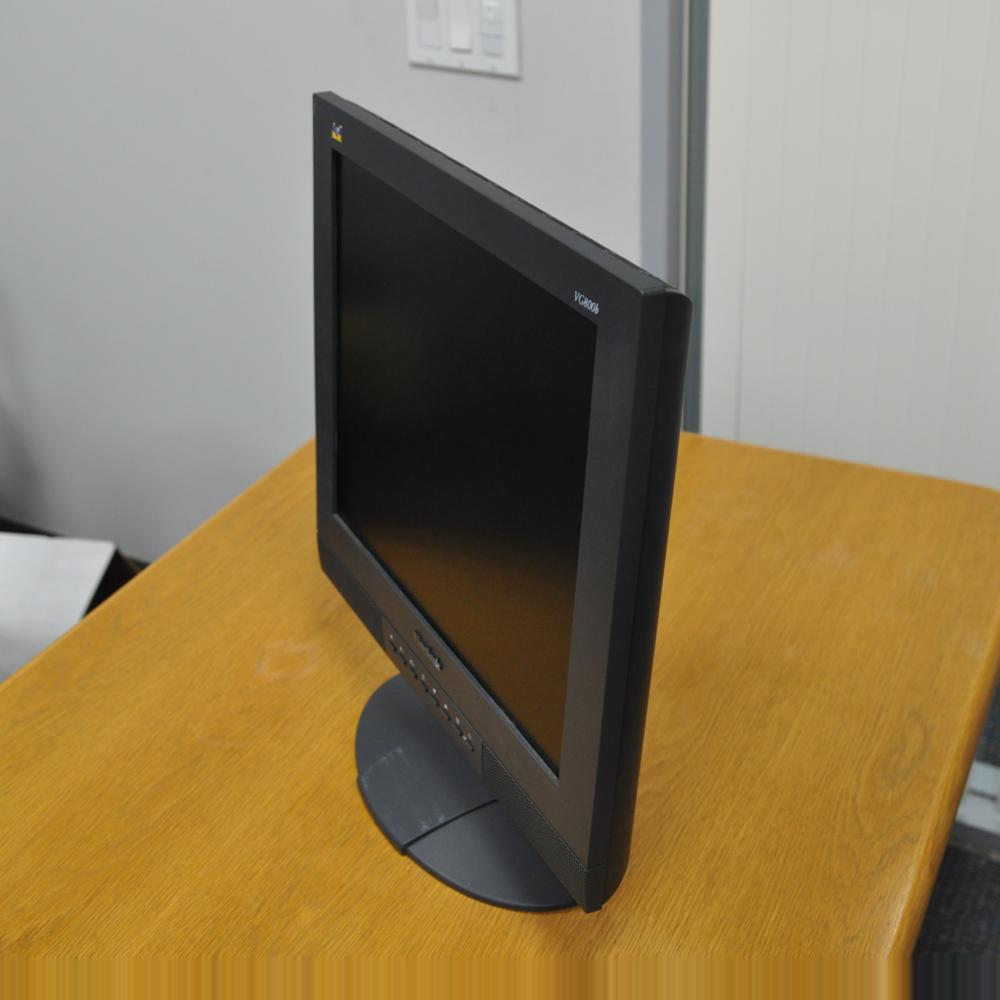}
       \includegraphics[width=0.3\linewidth,height=1.3cm]{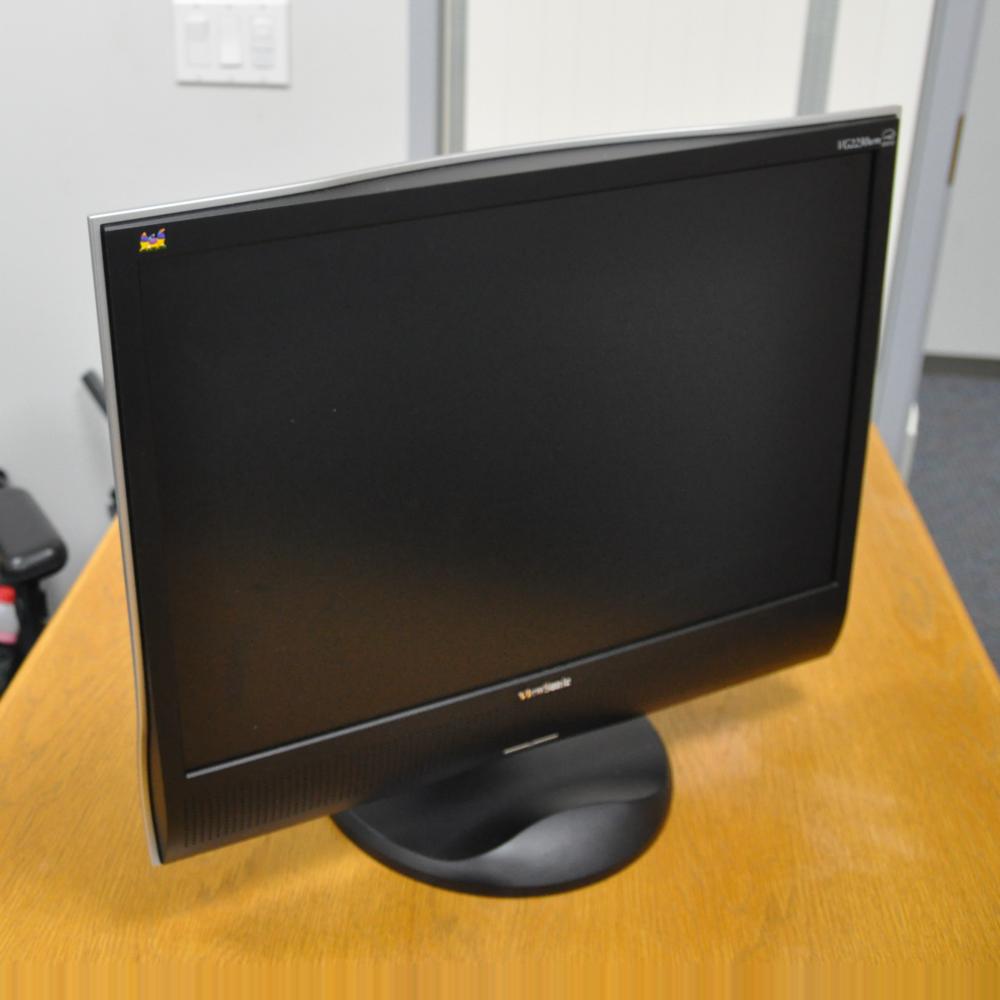}
        \includegraphics[width=0.3\linewidth,height=1.3cm]{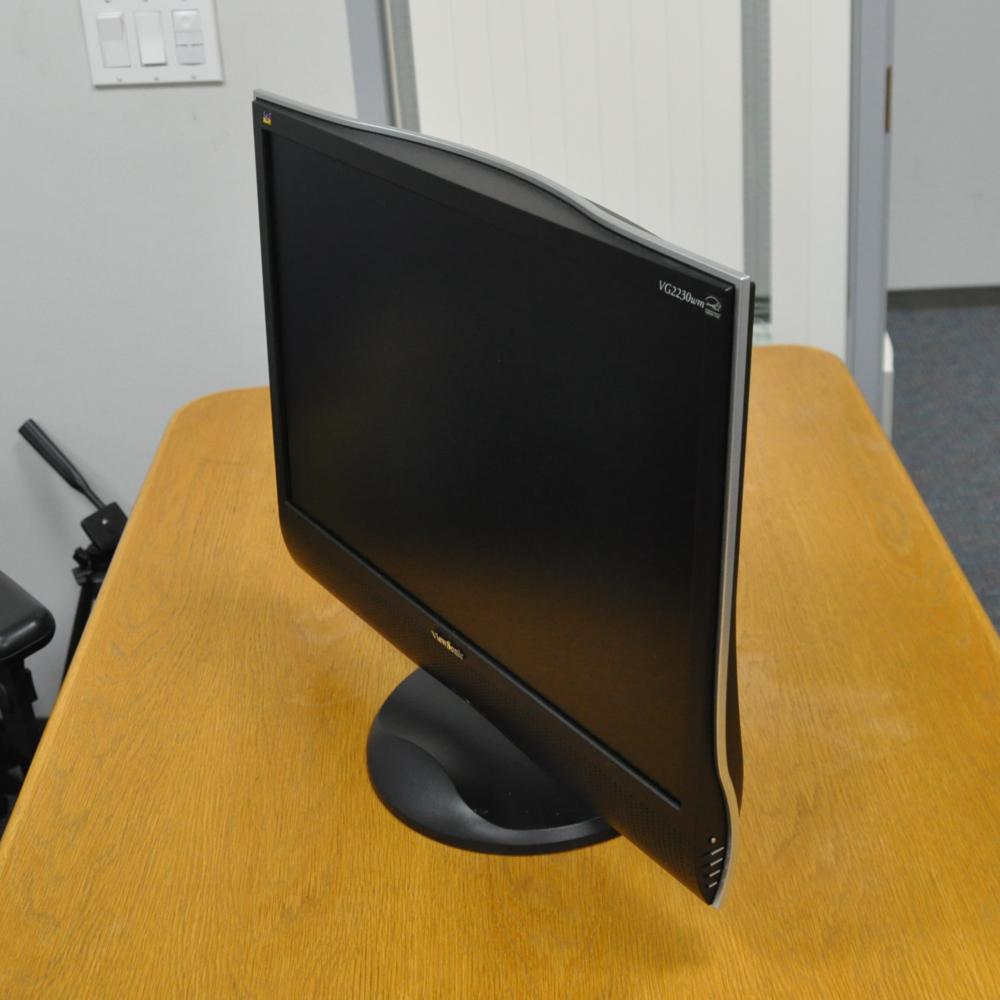}
         \includegraphics[width=0.3\linewidth,height=1.3cm]{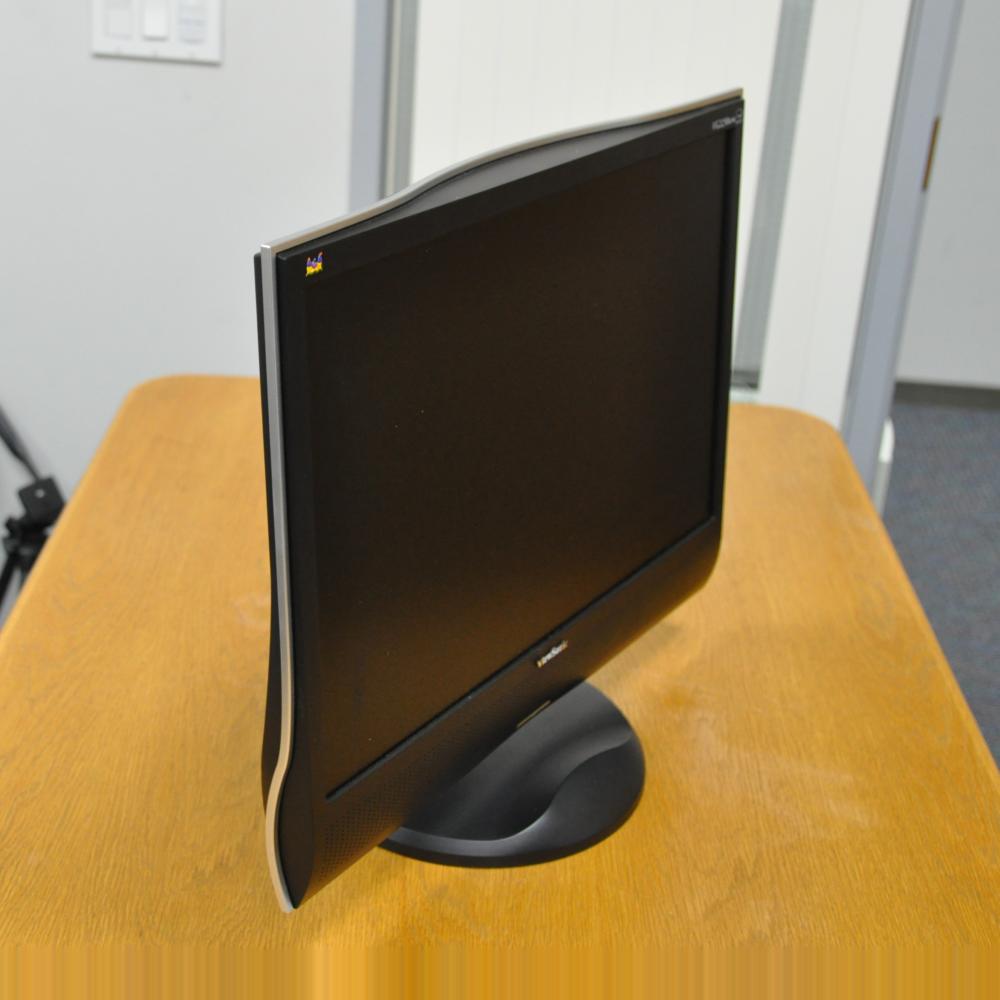}
    \caption*{DSLR}
    \label{fig:side:c}
  \end{minipage}
    \begin{minipage}[t]{0.49\linewidth}
   \centering
    \includegraphics[width=0.3\linewidth,height=1.3cm]{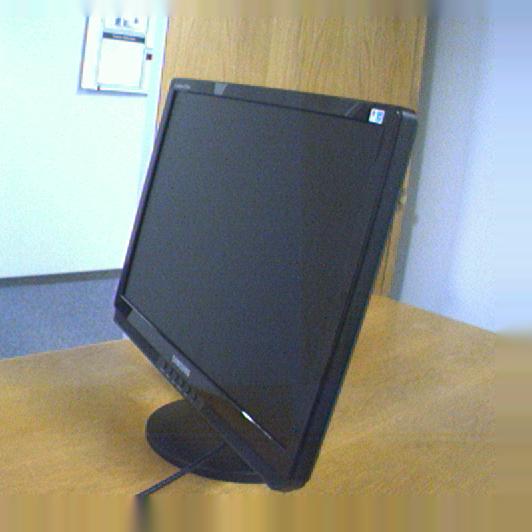}
     \includegraphics[width=0.3\linewidth,height=1.3cm]{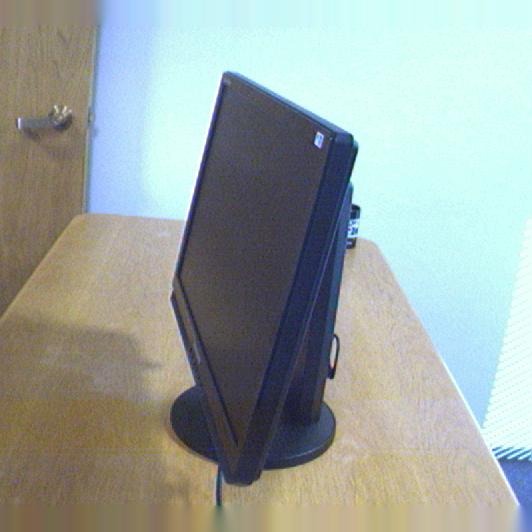}
      \includegraphics[width=0.3\linewidth,height=1.3cm]{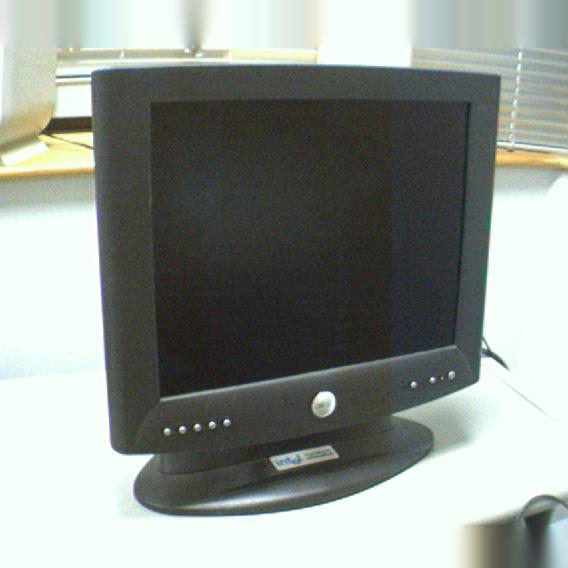}
       \includegraphics[width=0.3\linewidth,height=1.3cm]{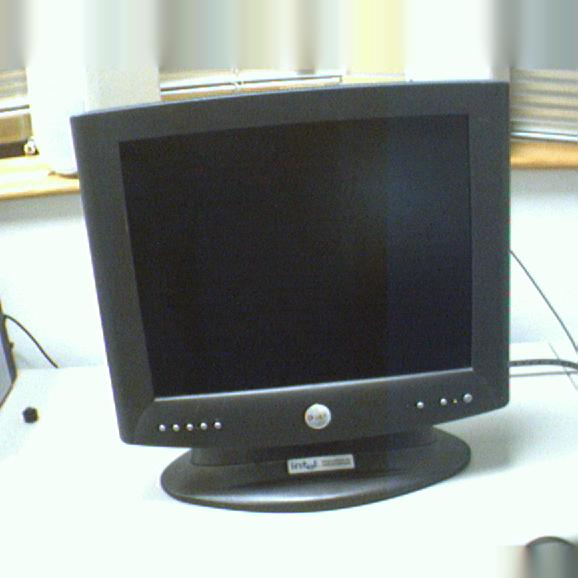}
        \includegraphics[width=0.3\linewidth,height=1.3cm]{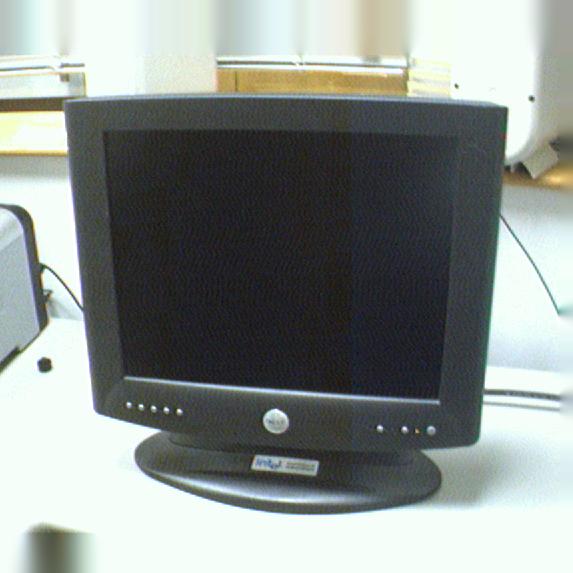}
         \includegraphics[width=0.3\linewidth,height=1.3cm]{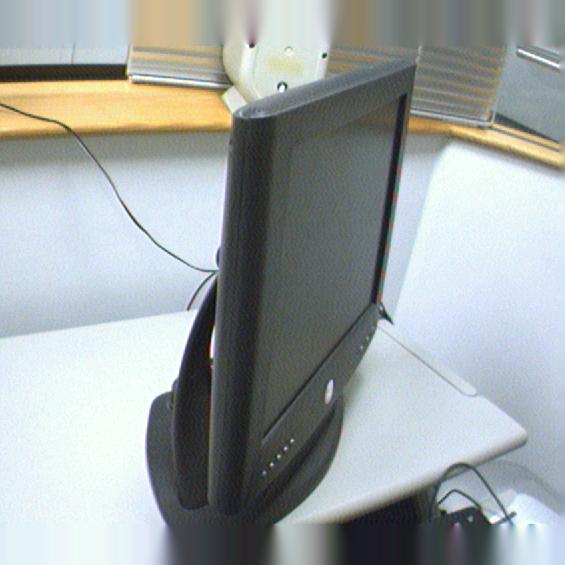}
    \caption*{Webcam}
    \label{fig:side:d}
  \end{minipage}
  \captionsetup{justification=centering}
   \caption{Some samples on 4DA office datasets}
   \label{4DAfig}
\end{figure}
\begin{figure}[t]
\centering
  \includegraphics[width=0.18\linewidth,height=1.3cm]{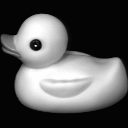}
  \includegraphics[width=0.18\linewidth,height=1.3cm]{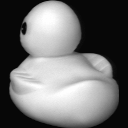}
  \includegraphics[width=0.18\linewidth,height=1.3cm]{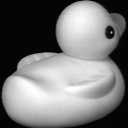}
  \includegraphics[width=0.18\linewidth,height=1.3cm]{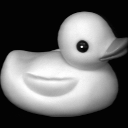}
  \includegraphics[width=0.18\linewidth,height=1.3cm]{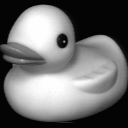}
  \includegraphics[width=0.18\linewidth,height=1.3cm]{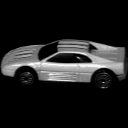}
  \includegraphics[width=0.18\linewidth,height=1.3cm]{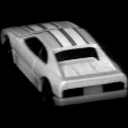}
  \includegraphics[width=0.18\linewidth,height=1.3cm]{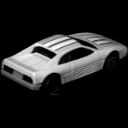}
  \includegraphics[width=0.18\linewidth,height=1.3cm]{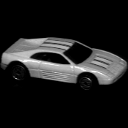}
  \includegraphics[width=0.18\linewidth,height=1.3cm]{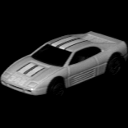}
  \includegraphics[width=0.18\linewidth,height=1.3cm]{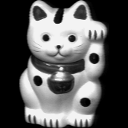}
  \includegraphics[width=0.18\linewidth,height=1.3cm]{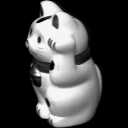}
  \includegraphics[width=0.18\linewidth,height=1.3cm]{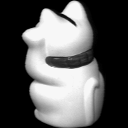}
  \includegraphics[width=0.18\linewidth,height=1.3cm]{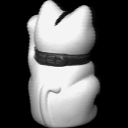}
  \includegraphics[width=0.18\linewidth,height=1.3cm]{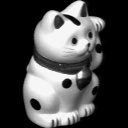}
  \includegraphics[width=0.18\linewidth,height=1.3cm]{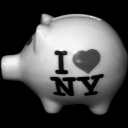}
  \includegraphics[width=0.18\linewidth,height=1.3cm]{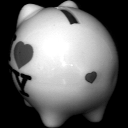}
  \includegraphics[width=0.18\linewidth,height=1.3cm]{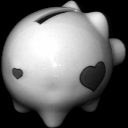}
  \includegraphics[width=0.18\linewidth,height=1.3cm]{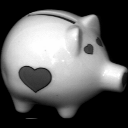}
  \includegraphics[width=0.18\linewidth,height=1.3cm]{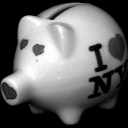}
\captionsetup{justification=centering}
   \caption{Some samples from COIL-20 dataset}
   \label{coil20}
\end{figure}
\begin{figure}[t]
  \begin{minipage}[t]{0.49\linewidth} % 如果一行放2个图，用0.5，如果3个图，用0.33
   \centering
    \includegraphics[width=0.3\linewidth]{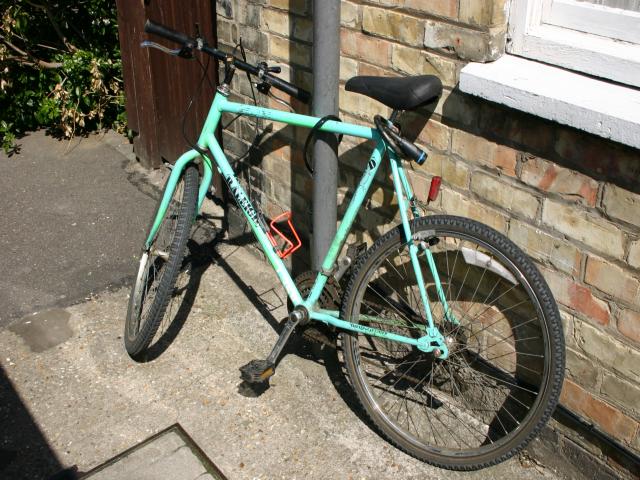}
     \includegraphics[width=0.3\linewidth]{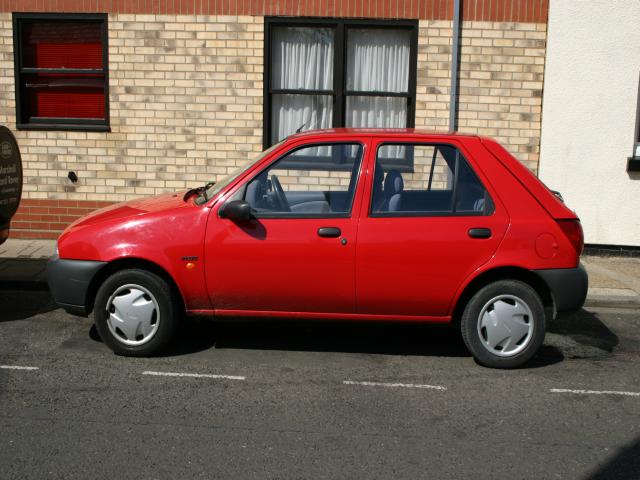}
      \includegraphics[width=0.3\linewidth]{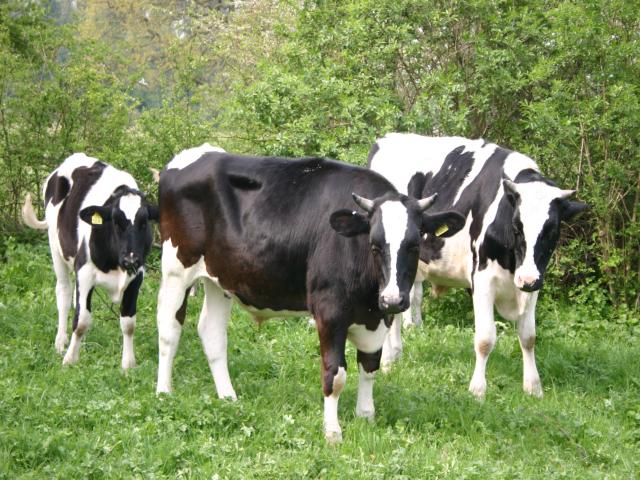}
       \includegraphics[width=0.3\linewidth]{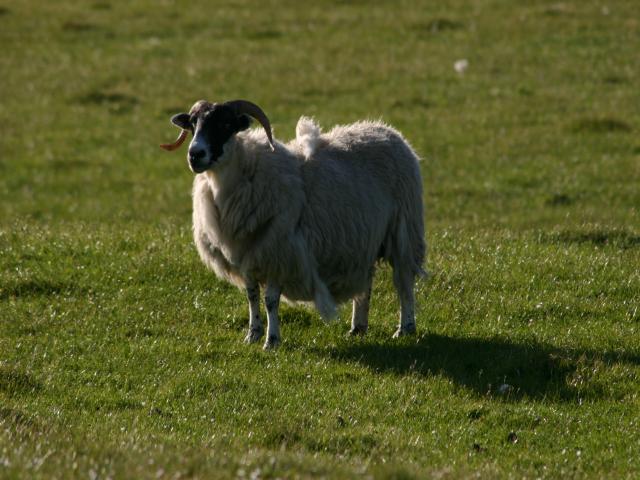}
        \includegraphics[width=0.3\linewidth]{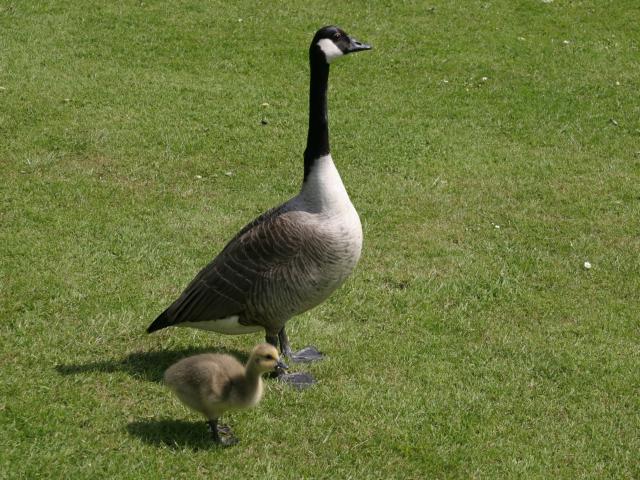}
         \includegraphics[width=0.3\linewidth]{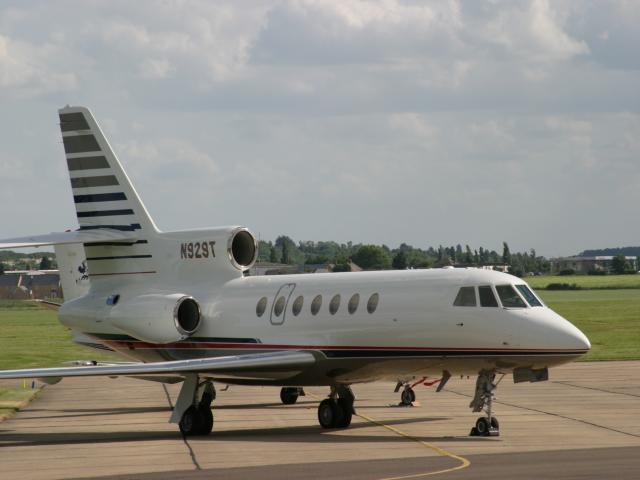}
  \caption*{ MSRC }
   \label{fig:side:a}
  \end{minipage}
  \begin{minipage}[t]{0.49\linewidth}
   \centering
    \includegraphics[width=0.3\linewidth]{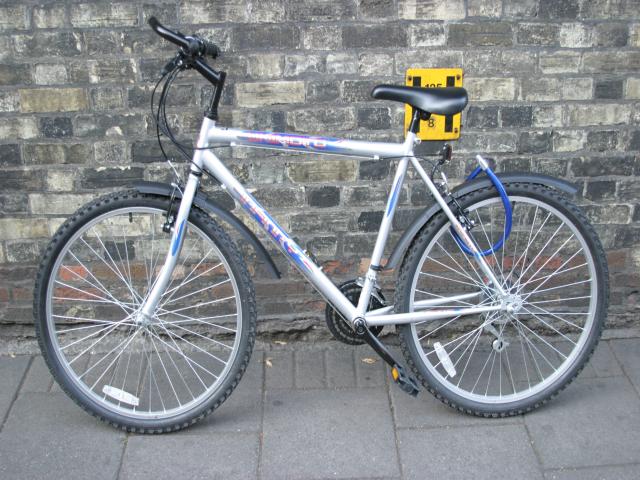}
     \includegraphics[width=0.3\linewidth]{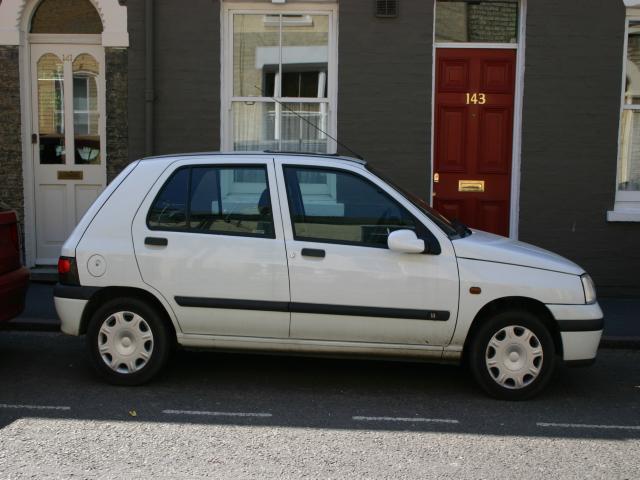}
      \includegraphics[width=0.3\linewidth]{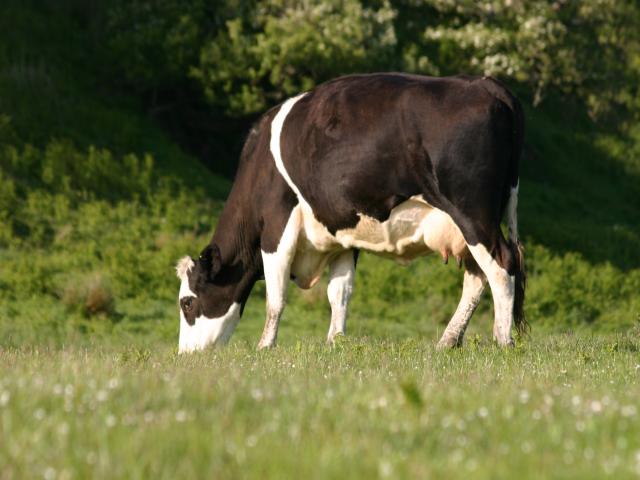}
       \includegraphics[width=0.3\linewidth]{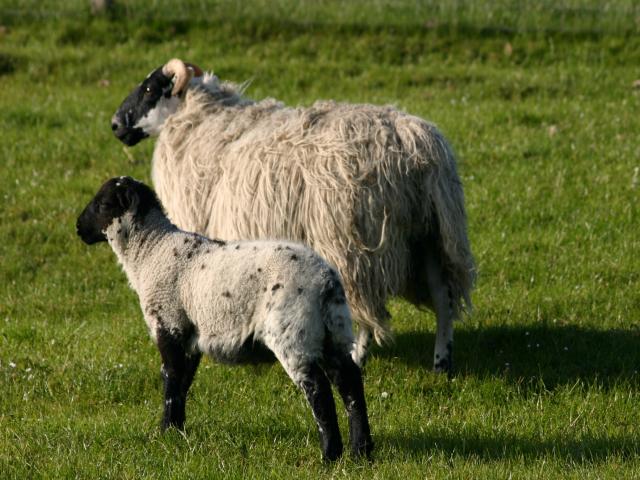}
        \includegraphics[width=0.3\linewidth]{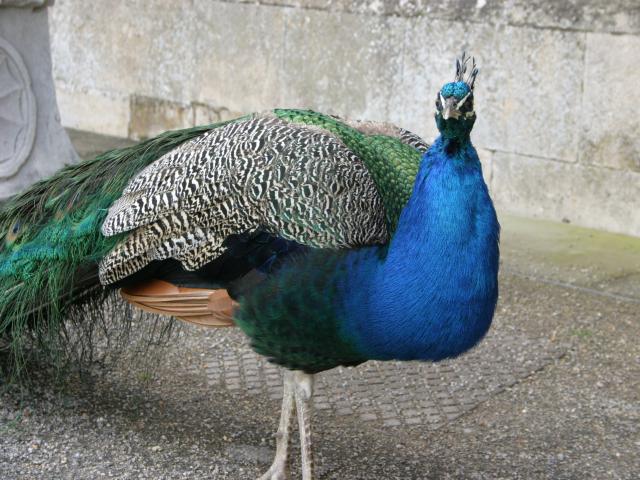}
         \includegraphics[width=0.3\linewidth]{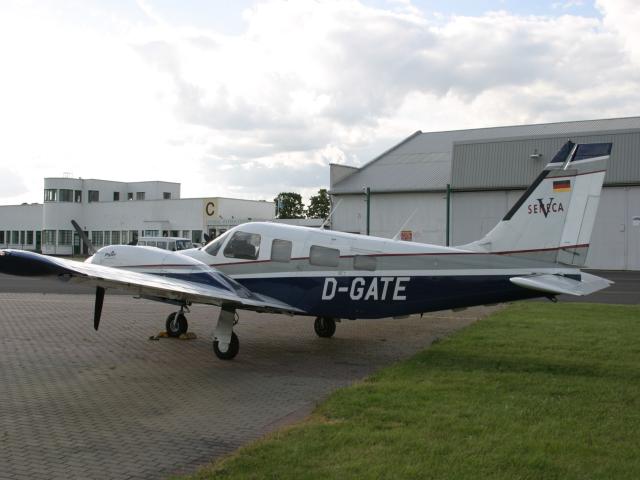}
    \caption*{VOC 2007}
    \label{fig:side:b}
  \end{minipage}
  \captionsetup{justification=centering}
   \caption{Some samples from MSRC and VOC 2007 datasets}
   \label{m2v}
\end{figure}
\begin{figure} [t]
\centering
  \includegraphics[width=1\linewidth]{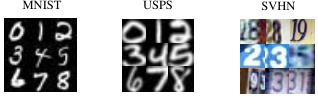}
\captionsetup{justification=centering}
   \caption{ Some examples from handwritten digits datasets}
   \label{fig3}
\end{figure}

{In this section, five benchmark datasets including 1 4DA office dataset, 2 office-31 dataset, 3 COIL-20 object dataset, 4 MSRC-VOC 2007 datasets and 5 cross-domain handwritten digits are conducted for cross-domain visual recognition.} Visually, in Fig. \ref{4DAfig}, it is shown some samples from 4DA office dataset. Several example images from COIL-20 object dataset are shown in Fig.~\ref{coil20}, several example images from MSRC and VOC 2007 datasets are described in Fig.~\ref{m2v}, and several example images from the handwritten digits datasets are shown in Fig.~\ref{fig3}. From the cross-domain images, the visual heterogeneity is significantly observed, and multiple cross-domain tasks are naturally formulated.

{ \textbf{Office 4DA (Amazon, Webcam, DSLR and Caltech 256) datasets }\cite{Gong2012Geodesic}:

There are four domains in Office 4DA datasets, which are Amazon (A), Webcam (W)\footnote{\url{http://www.eecs.berkeley.edu/~mfritz/domainadaptation/}}, DSLR (D) and Caltech (C)\footnote{\url{http://www.vision.caltech.edu/Image_Datasets/Caltech256/}}. These datasets contain 10 object classes. It is a common dataset in cross-domain problem. The experimental configuration in our method is followed in\cite{Gong2012Geodesic}, and the experiment extracted the 800-bin SURF features~\cite{Gong2012Geodesic} for comparison. Every two domains are selected as the source and target domain in turn.
As the source domain, 20 samples per object are selected from Amazon, and 8 samples per class are chosen from Webcam, DSLR and Caltech. However, as target training samples, 3 images per class are selected, while the rest samples in target domains are used as target testing data.  The experimental results of different domain adaptation methods are shown in Table~\ref{4DA}. From the results, we observe that the proposed CatDA  shows competitive recognition performance and the superiority is therefore proved.
Noteworthily, we also exhibit the unsupervised version of our method on this dataset. The asterisk (${^*}$) in Table~\ref{4DA} indicates that we use our method as an unsupervised manner and the results are not good. From the last lines in Table~\ref{4DA}, it is obvious that the semi-supervised manner is necessary and competitive in our simple network.}

%The detail is presented in Table \ref{tab1}.

\begin{table*}
\captionsetup{justification=centering}
\caption{Recognition performance ($\%$) of different transfer learning methods in 4DA office datasets}
\begin{center}
\setlength{\tabcolsep}{0.5mm}{
\begin{tabular}{  c | c  c  c  c c c  c c c  c | c c }
\hline
4DA Tasks&{NA}& HFA\cite{duan2012learning}&ARC-t\cite{kulis2011you}&MMDT\cite{Hoffman2014Asymmetric}&SGF\cite{Gopalan2011Domain}
&GFK\cite{Gong2012Geodesic}&SA\cite{Fernando2014Unsupervised}&\tabincell{c}{LTSL\\-PCA}\cite{shao2014generalized}&\tabincell{c}{LTSL\\-LDA}\cite{shao2014generalized}
&LSDT\cite{zhang2016lsdt}&$\bf{Ours^{*}}$&$\bf{Ours}$ \\
\hline
 $A \to D$&$55.9$&$52.7$&$50.2$&$56.7$&$46.9$&$50.9$&$55.1$&$50.4$&$59.1$&$52.9$&46.0&$\bf{60.5}$\\

 $C \to D$&$55.8$&$51.9$&$50.6$&$56.5$&$50.2$&$55.0$&$56.6$&$49.5$&$\bf{59.6}$&$56.0$&50.6&$59.4$\\

$W \to D$&$55.1$&$51.7$&$71.3$&$67.0$&$78.6$&$75.0$&$82.3$&$\bf{82.6}$&$\bf{82.6}$&$75.7$&68.3&$60.7$\\

 $A \to C$&$32.0$&$31.1$&$37.0$&$36.4$&$37.5$&$39.6$&$38.4$&$41.5$&$39.8$&$\bf{42.2}$&39.0&$37.6$\\

 $W \to C$&$30.4$&$29.4$&$31.9$&$32.2$&$32.9$&$32.8$&$34.1$&$36.7$&$\bf{38.5}$&$36.9$&33.0&$34.3$\\

 $D \to C$&$31.7$&$31.0$&$33.5$&$34.1$&$32.9$&$33.9$&$35.8$&$36.2$&$36.7$&$\bf{37.6}$&31.1&$36.9$\\

$D \to A$&$45.7$&$45.8$&$42.5$&$46.9$&$44.9$&$46.2$&$45.8$&$45.7$&$47.4$&$46.6$&39.3&$\bf{50.9}$\\

$W \to A$&$45.6$&$45.9$&$43.4$&$47.7$&$43.0$&$46.2$&$44.8$&$41.9$&$47.8$&$46.6$&41.5&$\bf{53.2}$\\

$C \to A$&$45.3$&$45.5$&$44.1$&$49.4$&$42.0$&$46.1$&$45.3$&$49.3$&$\bf{50.4}$&$47.7$&44.0&$\bf{50.4}$\\

$C \to W$&$60.3$&$60.5$&$55.9$&$63.8$&$54.2$&$57.0$&$60.7$&$50.4$&$59.5$&$57.6$&49.5&$\bf{67.6}$\\

$D \to W$&$62.1$&$62.1$&$78.3$&$74.1$&$78.6$&$80.2$&$\bf{84.8}$&$81.0$&$78.3$&$83.1$&72.7&$69.0$\\

$A \to W$&$62.4$&$61.8$&$55.7$&$64.6$&$54.2$&$56.9$&$60.3$&$52.3$&$59.5$&$57.2$&47.4&$\bf{68.0}$\\
\hline
$Average$&$48.5$&$47.4$&$49.5$&$52.5$&$49.7$&$51.6$&$53.7$&$51.5$&$\bf{54.9}$&$53.3$&46.9&$54.0$\\
\hline
\end{tabular}}
%\end{tabularx}
 \begin{tablenotes}\footnotesize
  \item  ~~~~~~~~~~~~The asterisk (${^*}$) indicates that we use our method as an unsupervised manner and therefore the results are not good.
 \end{tablenotes}
\end{center}
\label{4DA}
\end{table*}

\textbf{Office 3DA~(Office-31 dataset~\cite{saenko2010adapting})}:

{This dataset contains three domains such as Amazon (A), Webcam (W) and Dslr (D). It contains 4,652 images from 31 object classes. With each domain worked as source and target alternatively, 6 cross-domain tasks are formed, \textit{e.g.}, $A\to D$, $W\to D$, \textit{etc}. In experiment, we follow the experimental protocol as~\cite{GopalanUnsupervised} for the semi-supervised strategy.
In our method, 3 images per class are selected when they are used as target training data, while the rest samples in target domains are used for testing. The recognition accuracy is reported in Table~\ref{tab3DAshallow}. We can achieve the competitive results with~\cite{GopalanUnsupervised}. Noteworthily, we do not use any auxiliary and discriminative methods only the simple MLP network.
\begin{table*}[t]

\captionsetup{justification=centering}
\caption{Recognition accuracy ($\%$) of different transfer learning methods in office-31 dataset}
\begin{center}
\begin{tabular}{  c | c  c  c  c  c c  c }
\hline
Tasks&AVC\cite{saenko2010adapting}&AKT\cite{kulis2011you}&SGF\cite{Gopalan2011Domain}&GFK\cite{Gong2012Geodesic}&GMDA\cite{Zheng2012A}&GIDR\cite{GopalanUnsupervised}&\bf{Ours}\\
\hline
 A $\to$ W&$48.0$&$50.4$&$57.0$&$46.4$&$45.0$&$52.4$&$\bf{54.8}$\\
%\hline
D $\to$ W&$31.0$&$36.1$&$36.0$&$61.3$&$61.0$&$\bf{57.8}$&$57.2$\\
W $\to$ D&$27.0$&$25.3$&$37.0$&$66.3$&$66.0$&$\bf{58.3}$&$58.1$\\
\hline
$Average$&$35.3$&$37.3$&$43.3$&$\bf{58.0}$&$57.3$&$56.2$&$56.7$\\
\hline
%\end{tabularx}
\end{tabular}
\end{center}
\label{tab3DAshallow}

\end{table*}
}

\textbf{ Columbia Object Image Library~(COIL-20) dataset~\cite{Rate2011Columbia}}.

The COIL-20 dataset\footnote{\url{http://www.cs.columbia.edu/CAVE/software/softlib/coil-20.php}} contains 20 objects with 72 multi-pose images per class and total 1440 gray scale images. We follow the experimental protocol in~\cite{Xu2015}in our experiments, so this dataset is divided into two subsets C1 and C2, with each 2 quadrants are included. Specifically, the poses of quadrants 1 and 3 are selected as the C1 set and the C2 set contains the poses of quadrants 2 and 4. The two subsets are with different distribution due to the pose rotation but relevant in semantic describing the same objects, therefore it comes to a DA problem.
The subsets of C1 and C2 are chosen as source and target domain alternatively, and the cross-domain recognition performance of different methods are shown in Table~\ref{tab1}. From Table~\ref{tab1}, we can see that the proposed CatDA shows a significantly superior recognition performance ($93.0\%$) in average over other state-of-the-art shallow DA methods. This demonstrates that the proposed CatDA can generate similar feature representation with the target domain, such that the heterogeneity can be effectively reduced.

\begin{table*}[t]
\captionsetup{justification=centering}

\caption{Recognition performance ($\%$) of different transfer learning methods in COIL-20 dataset}
%\newcolumntype{Y}{>{\centering\arraybackslash}X}
%\begin{tabularx}{\textwidth}{|c|*{8}{Y|}}\hline
\begin{center}
\begin{tabular}{ c | c  c  c  c  c c  c }
\hline
%Tasks&SVM&TSL&RDALR\cite{Chang2013Robust}&LTSL\cite{shao2014generalized}&DTSL\cite{Xu2015}&LSDT\cite{zhang2016lsdt}&$\bf{Ours}$\\
Tasks&NA&TSL&RDALR\cite{Chang2013Robust}&LTSL\cite{shao2014generalized}&DTSL\cite{Xu2015}&LSDT\cite{zhang2016lsdt}&$\bf{Ours}$\\
\hline
$ C1 \to C2$&$82.7$&$80.0$&$80.7$&$75.4$&$84.6$&$89.7$&$\bf{94.3}$\\
%\hline
$ C2 \to C1$&$84.0$&$75.6$&$78.8$&$72.2$&$84.2$&$85.5$&$\bf{91.7}$\\
\hline
$Average$&$83.3$&$77.8$&$79.7$&$73.8$&$84.4$&$81.6$&$\bf{93.0}$\\
\hline
%\end{tabularx}
\end{tabular}
\end{center}
\label{tab1}

\end{table*}

\textbf{MSRC\footnote{\url{http://research.microsoft.com/en-us/projects/objectclassrecognition}} and VOC 2007\footnote{\url{http://pascallin.ecs.soton.ac.uk/challenges/VOC/voc2007}} datasets\cite{Xu2015}}:

The MSRC dataset contains 18 classes including 4323 images, and the VOC 2007 dataset contains 20 concepts with 5011 images.
These two datasets share 6 semantic categories: airplane, bicycle, bird, car, cow and sheep. In this way, the two domain data are constructed to share the same label set. The cross-domain experimental protocol is followed in~\cite{long2014transfer}. We select 1269 images from MSRC as the source domain and 1530 images from VOC 2007 as the target domain to construct a cross-domain task MSRC vs. VOC 2007. Then we switch the two datasets ( VOC 2007 vs. MSRC ) to  construct the other task.
For feature extraction, all images are uniformly re-scaled to 256 pixels, and the VLFeat open source package is used to extract the 128-dimensional dense SIFT (DSIFT) features. Then $K$-means clustering is leveraged to obtain a 240-dimensional codebook.

By following~\cite{WeiWangAAAI2017}, the source training sample set contains all the labeled samples in the source domain, 4 labeled target samples per class randomly selected from the target domain formulate the labeled target training data, and the rest unlabeled examples are recognized as the target testing data. The experimental results of different domain adaptation methods are shown in Table~\ref{M2V}. From the results, we observe that the proposed CatDA outperforms other DA methods.

\begin{table*}[t]

\captionsetup{justification=centering}
\caption{Recognition accuracy ($\%$) of different transfer learning methods in MSRC and VOC 2007}
\begin{center}
\begin{tabular}{  c | c  c  c  c  c c  c }
\hline
Tasks&NA&DTMKT-f\cite{Duan2012Domain}&MMDT\cite{Hoffman2014Asymmetric}&KMM\cite{Huang2006Correcting}&GFK\cite{Gong2012Geodesic}&LSDT\cite{zhang2016lsdt}&\bf{Ours}\\
\hline
 MSRC $\to$ VOC2007&$36.3$&$\bf{36.9}$&$36.0$&$36.1$&$29.5$&$\bf{36.9}$&$36.8$\\
%\hline
 VOC2007 $\to$ MSRC&$64.3$&$65.0$&$62.1$&$64.8$&$50.7$&$59.3$&$\bf{65.1}$\\
\hline
$Average$&$50.3$&$\bf{51.0}$&$49.1$&$50.5$&$40.1$&$48.1$&$\bf{51.0}$\\
\hline
%\end{tabularx}
\end{tabular}
\end{center}
\label{M2V}

\end{table*}

\begin{table*}[t]%\footnotesize

\captionsetup{justification=centering}
\caption{Recognition performance ($\%$) of shallow transfer learning in handwritten digits recognition}
%\newcolumntype{Y}{>{\centering\arraybackslash}X}
%\begin{tabularx}{\textwidth}{|c|*{9}{Y|}}\hline
\begin{center}
\begin{tabular}{ c | c  c  c  c  c c  c c }
\hline
 Tasks&{NA}&{A-SVM}& SGF\cite{Gopalan2011Domain}&GFK\cite{Gong2012Geodesic}&SA\cite{Fernando2014Unsupervised}&LTSL\cite{shao2014generalized}&LSDT\cite{zhang2016lsdt}&$\bf{Ours}$ \\
\hline
 MNIST $\to$ USPS&78.8&78.3&79.2&82.6&78.8&$\textbf{83.2}$&79.3&81.0\\
%\hline
 SEMEION $\to$ USPS&83.6&76.8&77.5&82.7&82.5&83.6&$\textbf{84.7}$&80.2\\
%\hline
MNIST $\to$ SEMEION&51.9&70.5&51.6&70.5&\bf{74.4}&72.8&69.1&$69.6$\\
%\hline
 USPS $\to$ SEMEION&65.3&74.5&70.9&$76.7$&74.6&65.3&67.4&\bf{78.0}\\
%\hline
 USPS $\to$ MNIST&71.7&73.2&71.1&$74.9$&72.9&71.7&70.5&\bf{75.6}\\
%\hline
 SEMEION $\to$ MNIST&67.6&69.3&66.9&$74.5$&72.9&67.6&70.0&\bf{75.8}\\
 \hline
$ Average$&69.8&73.8&69.5&\bf{77.0}&76.0&74.0&73.5&$76.7$\\
\hline
%\end{tabularx}
\end{tabular}
\end{center}
\label{tab2}
\end{table*}

\textbf{Cross-domain handwritten digits datasets}:

Three handwritten digits datasets: MNIST\footnote{\url{http://yann.lecun.com/exdb/mnist/}}, USPS\footnote{\url{http://www-i6.informatik.rwth-aachen.de/~keysers/usps.html}} and SEMEION\footnote{\url{http://archive.ics.uci.edu/ml/datasets/Semeion+Handwritten+Digit}} with 10 classes from digit $0\sim9$ are used for evaluating the proposed CatDA method. The MNIST dataset consists of 70,000 instances with image size of $28\times28$, the USPS dataset consists of 9,298 samples with image size of $16\times16$, and SEMEION dataset consists of 2593 images with image size of $16\times16$. In experiments, we adopt the strategy that crop the MNIST dataset into $16\times16$.
For DA experiment, every dataset of handwritten digits is used as the source domain and target domain alternatively, and totally 6 cross-domain tasks are obtained.
In experiment, 100 samples per class from source domain and 10 samples per category from target domain are randomly chosen for training. After the 5 random splits, the average classification accuracies are described in Table~\ref{tab2}. From the Table~\ref{tab2}, we observe that our method shows competitive recognition performance compared to other state-of-the-art methods in average, but slightly lower than the GFK classifier.%The superiority is therefore proved.

\subsection{Comparison with Some Deep Transfer Learning.}
%As the network can  be fed into deep feature,

{
In this section, some experiments are deployed on office-31 dataset and handwritten digit datasets for comparison with  state-of-the-art deep transfer learning approaches.

\textbf{Deep features of Office-31 dataset~\cite{saenko2010adapting}}:

%This dataset contains three domains such as Amazon (A), Webcam (W) and Dslr (D). It contains 4,652 images from 31 object classes. With each domain worked as source and target alternatively, 6 cross-domain tasks are formed, \textit{e.g.}, $A\to D$, $W\to D$, \textit{etc}. In experiment, we follow the experimental protocol as~\cite{zhang2016lsdt} for the semi-supervised strategy.
%In our method, 3 images per class are selected when they are used as target training data, while the rest samples in target domains are used for testing.
%The recognition accuracy is reported in Table~\ref{tab3DA}.
For fair comparison, we extract the deep features of Office-31 dataset by the ResNet-50 architecture. From the results, we observe that our method ($91.3\%$ in average) outperforms state-of-the-art deep domain adaptation methods. Especially, compared with ADGANet~\cite{Sankaranarayanan2017Generate} which is proposed from the generative method, our accuracy exceeds it by 4.8\%.  This demonstrates that our model can effectively alleviate the model bias problem.
Similar with 4DA dataset, we also exhibit the unsupervised version of our method on this dataset. The asterisk (${^*}$) in Table~\ref{tab3DA} indicates that we use our method as an unsupervised manner and the results are not so good.}
\begin{table*}[!htb]
\captionsetup{justification=centering}
\caption{Recognition performance ($\%$) of deep transfer learning methods in Office-31 dataset}
%\newcolumntype{Y}{>{\centering\arraybackslash}X}
%\begin{tabularx}{\textwidth}{|c|*{12}{Y|}}\hline
\begin{center}
%\begin{tabular}{ c |p{1.5cm}<{\centering} p{1.5cm}<{\centering}  p{2cm}<{\centering}  c c c c c c c c c }
\setlength{\tabcolsep}{0.55mm}\begin{tabular}{ c |  c c c c c c c c c c | c c }
\hline
Tasks&SourceOnly&{TCA}\cite{pan2011domain}&{GFK}\cite{Gong2012Geodesic}&{DDC}\cite{tzeng2014deep}&{DAN}\cite{long2015learning}&{RTN}\cite{long2016unsupervised}&{DANN}\cite{ganin2017domain}
&{ADDA}\cite{Tzeng2017Adversarial}&{JAN}\cite{long2017deep}&ADGANet\cite{Sankaranarayanan2017Generate}&$\bf{Ours^{*}}$&$\bf{Ours}$\\
%Tasks&Source only&Gradient reversal&Domain confusion&CoGAN&DANN&ADDA&$\bf{Ours}$\\
\hline
$A \to W$&$68.4$&$72.7$&$72.8$&$75.6$&$80.5$&$84.5$&$82.0$&$86.2$&$85.4$&$89.5$&$87.9$&$\bf{93.7}$\\
%\hline
$D \to W$&$96.7$&$96.7$&$95.0$&$96.0$&$97.1$&$96.8$&$96.9$&$96.2$&$97.4$&$\bf{97.9}$&$91.1$&$94.9$\\
%\hline
$W \to D$&$99.3$&$99.6$&$98.2$&$98.2$&$99.6$&$99.4$&$99.1$&$98.4$&$\bf{99.8}$&$\bf{99.8}$&$91.4$&$96.0$\\
%\hline
$A \to D$&$68.9$&$74.1$&$74.5$&$76.5$&$78.6$&$77.5$&$79.4$&$77.8$&$84.7$&$87.7$&$90.8$&$\bf{96.1}$\\
%\hline
$D \to A$&$62.5$&$61.7$&$63.4$&$62.2$&$63.6$&$66.2$&$68.2$&$69.5$&$68.6$&$72.8$&$72.0$&$\bf{84.0}$\\
%\hline
$W \to A$&$60.7$&$60.9$&$61.0$&$61.5$&$62.8$&$64.8$&$67.4$&$68.9$&$70.0$&$71.4$&$74.1$&$\bf{82.8}$\\
\hline
$Average$&$76.1$&$77.6$&$77.5$&$78.3$&$80.4$&$81.5$&$82.2$&$82.9$&$84.3$&$86.5$&$84.5$&$\bf{91.3}$\\
\hline
%\end{tabularx}
\end{tabular}
 \begin{tablenotes}\footnotesize
  \item ~~~~The asterisk (${^*}$) indicates that we use our method as an unsupervised manner and therefore the results are not good.
 \end{tablenotes}
\end{center}
\label{tab3DA}
\end{table*}

{
\textbf{Deep features of handwritten digits datasets}:
In this section, a new handwritten digits dataset SVHN\footnote{\url{http://ufldl.stanford.edu/housenumbers/}} (see Fig.~\ref{fig3}) is introduced. During generation, the content information is easy to be changed by GAN without supervision~(e.g., $3 \to 8$ in generation), however, in our method, the semi-supervised strategy can help completely avoid such incorrect generation.

%Three handwritten digits datasets including MNIST, USPS and SVHN (S)\footnote{\url{http://ufldl.stanford.edu/housenumbers/}} as shown in Fig.\ref{fig3} are used.
%Three handwritten digits datasets including MNIST(M), USPS(U) and SVHN(S) as shown in Fig.\ref{fig7} are used.
 %with 10 classes from digit $0\sim9$ are used for evaluating the proposed CAT method.
We have experimentally validated our proposed method in MNIST, USPS and SVHN datasets. These datasets share 10 classes of digits.
In our method, the deep features of three datasets are extracted using the LeNet model provided in the Caffe source code package.}
For adaptation tasks between MNIST and USPS, the training protocol established in~\cite{long2014transfer} is followed, where 2000 images from MNIST and 1800 images from USPS are randomly sampled. {While for adaptation task between SVHN and MNIST, we use the full training sets for comparison against~\cite{ganin2017domain}. All the domain adaptation tasks are conducted by following the experimental protocol in~\cite{Tzeng2017Adversarial}.}
%: MNIST $\to$ USPS, USPS $\to$ MNIST, and SVHN $\to$ MNIST following the \cite{Tzeng2017Adversarial}.
The key difference between~\cite{Tzeng2017Adversarial} and our method lies in that the ADDA~\cite{Tzeng2017Adversarial} is convolutional neural network structured method while ours is multilayer perceptron structured method. In ADDA, an essential problem is that the generated samples may be randomly changed (e.g., $3 \to 8$ instead of $3 \to 3$). In our CatDA, this problem can be handled by establishing multiple class-wise models.
In our setting, for target training data, 10 samples per class from target domain are randomly selected and 5 random splits are considered totally. The average classification accuracies are reported in Table~\ref{tab3}. { From the cross-domain recognition results, we observe that our CatDA model outperforms most state-of-the-art methods with $1\%$ improvement in average, and only lower than the very recent ADGANet~\cite{Sankaranarayanan2017Generate} method.}
\begin{table*}[!htb]
\captionsetup{justification=centering}
\caption{Recognition performance ($\%$) of deep transfer learning methods in handwritten digits recognition}
%\newcolumntype{Y}{>{\centering\arraybackslash}X}
%\begin{tabularx}{\textwidth}{|c|*{8}{Y|}}\hline
\begin{center}
\begin{tabular}{ c |p{1.5cm}<{\centering} p{1.5cm}<{\centering}  p{2cm}<{\centering}  c c c c  }
\hline
Tasks&Source only&DANN\cite{ganin2017domain}&Domain confusion\cite{tzeng2015simultaneous}&CoGAN\cite{Liu2016Coupled}&ADDA\cite{Tzeng2017Adversarial}&ADGANet\cite{Sankaranarayanan2017Generate}&$\bf{Ours}$\\
%Tasks&Source only&Gradient reversal&Domain confusion&CoGAN&DANN&ADDA&$\bf{Ours}$\\
\hline
 MNIST $\to$ USPS&$75.2$&$77.1$&$79.1$&${91.2}$&$89.4$&$\bf{92.8}$&$87.9$\\
%\hline
 USPS $\to$ MNIST&$57.1$&$73.0$&$66.5$&$89.1$&$90.1$&$90.8$&$\bf{95.9}$\\
%\hline
 SVHN$\to$ MNIST&$60.1$&$73.9$&$68.1$&$-$&${76.0}$&$\bf{92.4}$&$75.1$\\
\hline
$Average$&$64.1$&$74.7$&$71.2$&$-$&$85.2$&$\bf{92.0}$&${86.1}$\\
\hline
%\end{tabularx}
\end{tabular}
\end{center}
\label{tab3}
\end{table*}

\section{Discussion}
\subsection{Computational Efficiency}
As the proposed CatDA model is simple in cross-domain visual recognition system, CPU is enough for model optimization and training instead of GPU. Also, the time cost is much lower as shown in Table~\ref{tab4}, from which we can observe that the computational speed is quite fast for three layered MLP model. Considering that our three
layered shallow network can achieve competitive performance and fast computational time, we choose this three layered model in our method.
Our experiments are implemented on the runtime environment with a PC of Intel i7-4790K CPU, 4.00GHz, 32GB RAM. It is noteworthy that the time for data preprocessing and classification is excluded.

%\begin{figure}[!htb]
%\begin{center}
%  \includegraphics[width=1\linewidth]{figure//GENvisualization2.pdf}
%\end{center}
%%\captionsetup{justification=centering}
%   \caption{(a) The generated image ${\hat{\bf{X}}_{ST}}$ (USPS\_generated) and ${{\bf{X}}_{T}}$ (USPS). (b) The generated image ${\hat{\bf{X}}_{TS}}$ (MNIST\_generated) and ${{\bf{X}}_{S}}$ (MNIST). (c) The first two illustrate the generated data ${\hat{\bf{X}}_{ST}}$ and ${\hat{\bf{X}}_{TS}}$ and the last two show the generated data ${\hat{\bf{X}}_{STS}}$ and ${\hat{\bf{X}}_{TST}}$.}
%   \label{fig4}
%\end{figure}

\subsection{Evaluation of Layer Number}

For insight of the impact of layers, we show the results of different number of layers in Table~\ref{tab5}, from which we can observe that the performance does not always show an upward trend with increasing layers and the three layered shallow network can achieve competitive performance.

\begin{table}[h]

\caption{The time cost ($s$) with different number of layers on different datasets.}
\begin{center}
\begin{tabular}{  c | c c c  c }
\hline
Tasks&3 layers& 4 layers&5 layers& 6 layers \\
\hline
$ C1 \to C2$&$17.25$&$51.09$&$102.48$&$148.02$\\
%\hline
%$ C2 \to C1$&$0.128$&$0.223$&$0.346$&$0.472$\\
%\hline
MSRC $\to$ VOC&$4.38$&$8.83$&$17.50$&$25.44$\\
%\hline
%VOC $\to$ MSRC&$0.298$&$1.098$&$2.095$&$3.171$\\
%\hline
SVHN$\to$ MNIST&$753.72$&$2275.78$&$4090.25$&$7600.42$\\
\hline
\end{tabular}
\end{center}
\label{tab4}

\end{table}

\begin{table}[h]
\caption{Recognition performance ($\%$) with different number of layers on  different datasets.}

\begin{center}
\begin{tabular}{  c | c c c  c }
\hline
Tasks&3 layers& 4 layers&5 layers& 6 layers \\
\hline
$ C1 \to C2$&$\bf{94.3}$&$93.3$&$93.3$&$93.7$\\
%\hline
$ C2 \to C1$&$91.7$&$\bf{91.8}$&$91.3$&$90.8$\\
\hline
 MNIST $\to$ USPS&$87.9$&$\bf{88.3}$&$88.2$&$\bf{88.3}$\\
%\hline
 USPS $\to$ MNIST&$95.9$&$96.0$&$\bf{96.1}$&$96.0$\\
%\hline
SVHN$\to$ MNIST&$75.1$&$84.8$&$91.1$&$\bf{92.8}$\\
\hline
Average&$89.0$&$90.9$&$92.0$&$\bf{92.3}$\\
\hline
\end{tabular}
\end{center}
\label{tab5}

\end{table}

\begin{figure*} [!htb]
\begin{center}
  \includegraphics[width=0.8\linewidth]{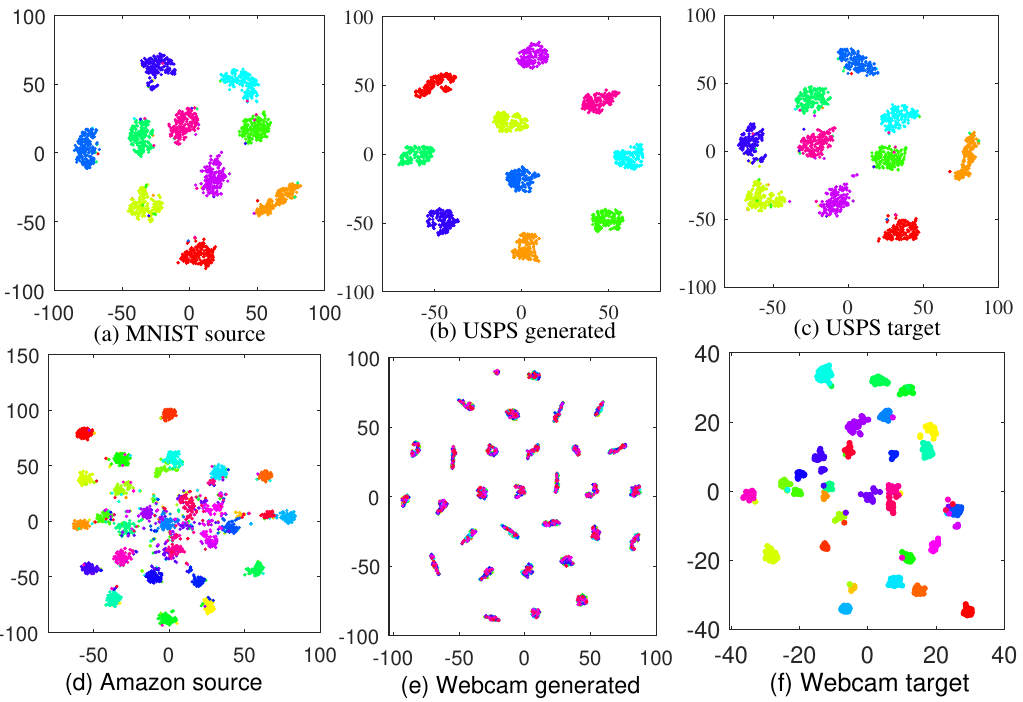}
\end{center}
%\captionsetup{justification=centering}
   \caption{Feature visualization of data distribution. Both (a) and (c) show the feature distribution of source domain (MNIST) and target domain (USPS). (b) shows the feature distribution of the generated data ($ M \to \hat{U}$).
   Both (d) and (f) show the feature distribution of source domain (Amazon) and target domain (Webcam). (e) shows the distribution of the generated samples ($ A \to \hat{W}$).}%It is obvious that the generated domain is closer with the target domain after projection from the range of abscissa axis.
   \label{fig5}
\end{figure*}

\subsection{Model Visualization}
{In this section, for better insight of the CatDA model, visualization of class distribution is explored. We visualize the features for further validating the effectiveness of our model.
The t-SNE~\cite{donahue2014decaf} visualization method  is employed on the source domain and target domain in the $ {M} \to {U}$ task of handwritten digits datasets~(shallow domain adaptation) and $ A \to W$ task of 3DA office datasets~(deep domain adaptation) for feature visualization.  From the (b) and (e) in Fig.~\ref{fig5}, it is obvious that the better clustering characteristic is achieved and the feature discriminative power is improved in the generated data. As a result, the visual recognition cross-domain performance is improved. It is worthy noting that as shown in Fig.~\ref{fig5} the discrimination of the generated features becomes better than raw features. The reason is that in our semi-supervised domain adaptation method, a partial target label information is used for feature generation, which, therefore, improves the discrimination of the generated features.}

\subsection{Remarks}
{The proposed method is an adversarial feature adaptation model, which can be used for semi-supervised and unsupervised domain adaptation. We have to claim that 1) the proposed CatDA is completely different from GAN that it cannot be used for image synthesis in its current form, but only for domain adaptive feature representation. This is because the inputs fed into our CatDA are still features such as the low-level hand-crafted feature or deep features by an off-the-shelf CNN model, rather than image pixels. 2) The proposed model is not CNN based, so it may not compare with deep networks. However, deep features can be fed into our model for fair comparison as is presented in Table~\ref{tab3DA}. }
% It is obvious that the generated domain is closer with the target domain after projection from the range of abscissa axis. Additionally,

%\begin{table}[t]\small
%\caption{Recognition accuracy ($\%$) of Class-wise CatDA and Conditional CatDA}
%\begin{center}
%\begin{tabular}{  c | c  c c }
%\hline
%Tasks&Class-wise CatDA& Conditional CatDA \\
%\hline
%$ M \to U$&$87.3$&$82.9$\\
%%\hline
%$ U \to M$&$95.5$&$93.7$\\
%%\hline
%$ S\to M$&$92.1$&$95.0$\\
%%\hline
%Average&$91.4$&$91.1$\\
%\hline
%\end{tabular}
%\end{center}
%\label{tab5}
%\end{table}

\section{Conclusion}
In this paper, we propose a new transfer learning method from the perspective of feature generation for cross-domain visual recognition. Specifically, a coupled adversarial transfer domain adaptation (CatDA) framework comprising of two generators, two discriminators, two domain specific loss terms and two content fidelity loss terms is proposed in a semi-supervised mode for domain and intra-class discrepancy reduction. This symmetric model can achieve bijective mapping, such that the domain feature can be generated alternatively benefiting from the reversible characteristic of the proposed model. Consider that our focus is the domain feature generation with distribution disparity removed for cross-domain applications, rather than realistic image generation, a shallow yet effective MLP transfer network is therefore considered. Extensive experiments on several benchmark datasets demonstrate the superiority of the proposed method over some other state-of-the-art DA methods.

In our future work, benefit from the strong feature learning capability of deep neural network (e.g. convolutional neural network), deep adversarial domain adaptation framework that owns similar model with CatDA should be focused. Compared with CatDA, deep methods extract high dimensional semantic features to achieve good performance but they need large-scale samples to train the network. Hence, fine-tuning based parameter transfer from big data to small data can be leveraged for improving the network adaptation.

\section*{Acknowledgment}The authors are grateful to the AE and anonymous reviewers for their valuable comments on our work.


\ifCLASSOPTIONcaptionsoff
  \newpage
\fi

{\small
\bibliographystyle{ieee}
\bibliography{catda18}
}

\begin{IEEEbiography}[{\includegraphics[width=1in,height=1.25in,clip,keepaspectratio]{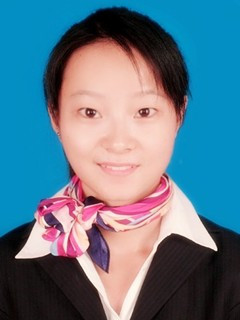}}]{Shanshan Wang}
 received BE and ME from the Chongqing University in 2010 and 2013, respectively. She is currently pursuing the Ph.D. degree at Chongqing University. Her current research interests include machine learning, pattern recognition, computer vision.
\end{IEEEbiography}

\begin{IEEEbiography}[{\includegraphics[width=1in,height=1.25in,clip,keepaspectratio]{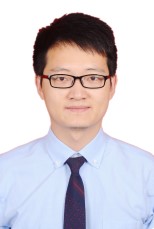}}]{Lei Zhang}
%\begin{IEEEbiography}{Lei Zhang}
(M'14-SM'18) received his Ph.D degree in Circuits and Systems from the College of Communication Engineering, Chongqing University, Chongqing, China, in 2013. He worked as a Post-Doctoral Fellow with The Hong Kong Polytechnic University, Hong Kong, from 2013 to 2015. He is currently a Professor/Distinguished Research Fellow with Chongqing University. He has authored more than 90 scientific papers in top journals, such as IEEE T-NNLS, IEEE T-IP, IEEE T-MM, IEEE T-IM, IEEE T-SMCA, and top conferences such as ICCV, AAAI, ACM MM, ACCV, etc. His current research interests include machine learning, pattern recognition, computer vision and intelligent systems. Dr. Zhang was a recipient of the Best Paper Award of CCBR2017, the Outstanding Reviewer Award of many journals such as Pattern Recognition, Neurocomputing, Information Sciences, etc., Outstanding Doctoral Dissertation Award of Chongqing, China, in 2015, Hong Kong Scholar Award in 2014, Academy Award for Youth Innovation in 2013 and the New Academic Researcher Award for Doctoral Candidates from the Ministry of Education, China, in 2012.
\end{IEEEbiography}

\begin{IEEEbiography}[{\includegraphics[width=1in,height=1.25in,clip,keepaspectratio]{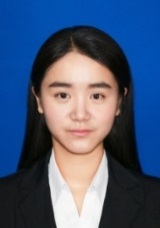}}]{Jingru Fu}
 received the B.S. degree from Fuzhou University, Fuzhou, China, in 2017. She is currently working towards the M.S. degree in Learning Intelligence and Vision Essential (LiVE) group at Chongqing University, Chongqing, China. Her current research interests include machine learning, transfer learning and computer vision.
\end{IEEEbiography}
%% insert where needed to balance the two columns on the last page with
%%\newpage


%\vfill


\end{document}